\documentclass[conference]{IEEEtran}
\IEEEoverridecommandlockouts
\usepackage{cite}
\usepackage{amsmath,amssymb,amsfonts}
\usepackage{graphicx}
\usepackage{textcomp}
\usepackage{xcolor}

\usepackage{comment}
\usepackage{algorithm}
\usepackage[noend]{algpseudocode}
\usepackage{amsmath}
\usepackage{amssymb}

\usepackage{url}
\usepackage{booktabs}

\def\BibTeX{{\rm B\kern-.05em{\sc i\kern-.025em b}\kern-.08em
    T\kern-.1667em\lower.7ex\hbox{E}\kern-.125emX}}
\begin{document}

\title{Learning 
Sparse Decision Trees via Transformer Variational Auto-Encoders}

\author{
\IEEEauthorblockN{Giacomo Fidone}
\IEEEauthorblockA{
\textit{University of Pisa}\\
giacomo.fidone@phd.unipi.it
}
\and
\IEEEauthorblockN{Alessio Cascione}
\IEEEauthorblockA{
\textit{University of Pisa}\\
alessio.cascione@phd.unipi.it
}
\and
\IEEEauthorblockN{Riccardo Guidotti}
\IEEEauthorblockA{
\textit{University of Pisa}\\
\textit{ISTI-CNR}\\
riccardo.guidotti@unipi.it
}
}

\newcommand\blfootnote[1]{%
  \begingroup
  \renewcommand\thefootnote{}\footnote{#1}%
  \addtocounter{footnote}{-1}%
  \endgroup
}

\newcommand{\trevis}[1]{{\textsc{trevis}}}
\newcommand{\ttvae}[1]{{\textsc{ttvae}}}

\maketitle

\begin{abstract}
Decision trees are among the most widely used models in machine learning, largely due to their transparent decision logic, making them well-suited for high-stakes decision-making contexts. 
However, most existing learning algorithms focus on predictive performance, overlooking the joint optimization of other desirable properties, such as structural sparsity.
In this work we propose TREVIS, an approach for learning decision trees with respect to complex objectives, based on the exploration of the latent space of a Tree Transformer Variational Auto-Encoder (TTVAE). 
By mapping decision trees onto latent representations, TREVIS replaces the discrete search space with a continuous one, enabling gradient-based optimization via a differentiable surrogate model.
We experiment with TREVIS for learning decision trees that jointly optimize predictive performance and sparsity. 
Results show that TREVIS discovers decision trees matching the predictive performance of existing near-optimal algorithms while improving their structural sparsity.
\end{abstract}


\section{Introduction}\blfootnote{\textbf{Accepted for publication at the 2026 IEEE International Conference on Data Mining (ICDM 2026).}
\textcopyright~2026 IEEE. Personal use of this material is permitted.
Permission from IEEE must be obtained for all other uses, in any current or future media, including reprinting/republishing this material for advertising or promotional purposes, creating new collective works, for resale or redistribution to servers or lists, or reuse of any copyrighted component of this work in other works.}
Machine Learning (ML)\blfootnote{This work has been partially supported by the Italian Project Fondo Italiano per la Scienza FIS00001966 “MIMOSA”, by the European Community programme under the funding schemes G.A. 101120763 ``TANGO'', and G.A. 101286379 “ILLUME-4-Science”.} models are increasingly deployed in high-stakes decision-making contexts such as credit risk assessment, hiring, and healthcare~\cite{bhatore2020machine,DeFauw2018ex1,garg2022review}.
Despite their strong predictive capabilities, many ML models rely on uninterpretable architectures compromising safety and accountability~\cite{bodria2023benchmarking}. 
Given their hierarchical, rule-based structure, Decision Trees (\textsc{dt}s)~\cite{breiman1984classification} stand out as one of the few interpretable-by-design models and remain a standard for tabular data~\cite{Rudin19stop}.

However, learning optimal \textsc{dt}s is intractable for the combinatorial size of the discrete search space, which grows exponentially with the number of features and training instances~\cite{laurent1976constructing}. Consequently, algorithms for learning \textsc{dt}s only explore a small region of this space.
The most common ones, namely CART, ID3, and C4.5~\cite{Rokach2005topdownlearnsurvey}, rely on a top-down greedy strategy that, while computationally efficient, typically yields suboptimal solutions. 
In contrast, algorithms seeking globally optimal \textsc{dt}s incur prohibitive computational costs~\cite{bertsimas2017optimal,aghaei2025strong}.

Beyond the performance-efficiency trade-off, most learning algorithms typically focus on optimizing predictive performance, with little or no account for other desirable properties~\cite{loh2014fifty}. 
Notably, the interpretability of a \textsc{dt} depends on its \textit{structural sparsity}, namely on how simple the tree topology is in terms of depth and the number of nodes and leaves~\cite{huysmans2011empirical}. 
Additionally, a \textsc{dt} may be required to ensure \textit{fairness} w.r.t. one or more protected attributes~\cite{Mehrabi2021fairnesssurv}, preserve \textit{privacy} by masking sensitive information from possible adversarial attacks~\cite{monreale2014privacy} or display \textit{robustness} to noisy input examples~\cite{wang2023trustworthy}. 
\begin{figure}[t]
    \centering
    \includegraphics[width=0.82\linewidth, trim={0cm 5.2cm 0cm 0cm}, clip]{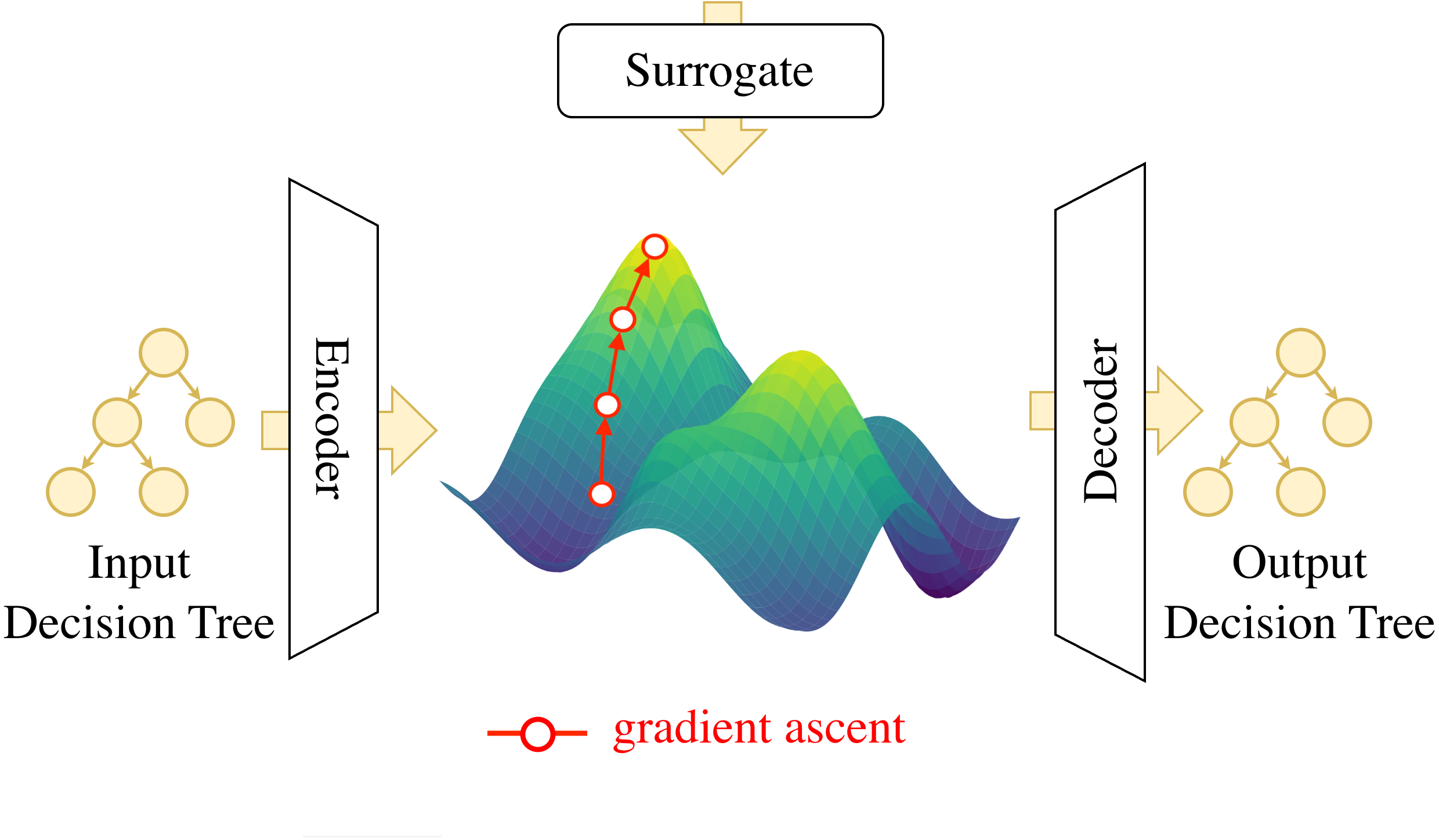}
    \caption{\trevis{} learns a latent space of Decision Trees via a Tree Transformer Variational Auto-Encoder. Then, it leverages gradient-ascent to find latent representations maximizing desired properties predicted by a surrogate model.}
    \label{fig:trevis}
\end{figure}
A promising alternative to existing \textsc{dt} learning algorithms is to embed the discrete space of \textsc{dt}s into a smoother, continuous latent space. 
This makes exploration more efficient and enables the optimization of complex objectives via black-box or gradient-based methods, as previously demonstrated for other graph-structured domains~\cite{zhang2019dvae,kusner2017grammar}.
Following this line of research, we propose a framework for learning \textsc{t}ree \textsc{re}presentations from \textsc{v}ariational \textsc{i}nference in latent \textsc{s}pace (\trevis{}). 
As shown in Figure~\ref{fig:trevis}, \trevis{} learns a continuous latent space of \textsc{dt}s via a \textsc{t}ree \textsc{t}ransformer \textsc{v}ariational \textsc{a}uto-\textsc{e}ncoder (\ttvae{}). 
Then, it leverages a differentiable surrogate model for jointly estimating desired properties  and discovering optimal latent tree representations with gradient ascent. 

In this work, we evaluate the ability of \trevis{} to generate \textsc{dt}s that balance two competing properties: predictive performance and structural sparsity. 
This trade-off is central to \textsc{dt} learning, as more complex \textsc{dt}s can improve performance, but often at the expense of interpretability. 
We leave to future work the extension of \trevis{} to more complex objectives including fairness, privacy and robustness.
Our results show that \trevis{} learns a locally smooth latent space with identifiable directions capturing \textsc{dt} properties. This enables \trevis{} to effectively navigate the latent space and find \textsc{dt}s with predictive performance comparable to that of near-optimal \textsc{dt} learning algorithms, while improving structural sparsity.

The rest of the paper is organized as follows. In Section~\ref{sec:related}, we review related works. In Section~\ref{sec:method}, we formalize our proposal. In Section~\ref{sec:experiments}, we report experimental setting and results. Finally, in Section~\ref{sec:conclusion}  we summarize our contributions and detail future research directions.

\section{Related Works}
\label{sec:related}
We review related works on \textsc{dt} learning algorithms, Latent Space Optimization (LSO) via Variational Auto-Encoders (VAEs) and transformer-based models for tree-structured data.

\smallskip
\textit{Decision Tree Learning.} Traditional methods for learning \textsc{dt}s rely on greedy top-down approaches, recursively partitioning data to minimize impurity measures~\cite{breiman1984classification,tan2006data}. 
While efficient, these approaches are prone to overfitting and attribute selection bias~\cite{hawkins2004problem,hothorn2006unbiased}. 
To address their limitations, optimal tree learning methods based on mathematical programming have been proposed, although their computational cost is often prohibitive~\cite{bertsimas2017optimal,vilas2021optimal}. 
As a compromise, recent work has explored methods for learning near-optimal \textsc{dt}s, including mathematical programming solvers, dynamic-programming and branch-and-bound techniques, and metaheuristic search~\cite{farhangfar2008fast,aglin2020learning,Demirovic2022otpimaldynamic,rivera2022induction}. 
Some of these approaches further control tree complexity through explicit structural regularization~\cite{Lin2020Generalized,mctavish2022fast}. 
In contrast, \trevis{} learns a continuous latent space in which \textsc{dt}s can be explored and optimized.
To the best of our knowledge, \cite{guidotti2024gentree} is the only prior generative approach to \textsc{dt} learning. However, it applies convolutions on a matrix encoding of \textsc{dt}s, which is less expressive for tree-structured data; and uses a sample-inefficient black-box optimization. In contrast, \trevis{} relies on a transformer-based VAE explicitly capturing tree structure and leverages efficient gradient-based optimization to discover optimal \textsc{dt}s in latent space. 

\smallskip
\textit{Latent Space Optimization.} VAEs~\cite{kingma2019introduction} are widely used for generative modeling and representation learning~\cite{pu2016variational,redah2026autoencoders}, due to their capability of learning structured latent spaces that can be navigated for controlled data generation. Building on this property, LSO optimizes latent representations based on a target objective by using black-box or gradient-based methods~\cite{biswas2023optimizing}. Recent works have explored the use of LSO on structured data. For example,~\cite{kusner2017grammar, jin2018junction} propose VAEs for learning latent vectors of molecules, enabling the discovery of new compounds with desired chemical properties. 
Related approaches have also been proposed for directed acyclic graphs~\cite{zhang2019dvae,thost2021directed}, targeting tasks such as neural architecture search and Bayesian network optimization.
LSO techniques range from gradient-based methods based on surrogate models, such as sparse Gaussian processes with expected improvement~\cite{kusner2017grammar,snelson2005sparse,jones1998efficient}, to black-box heuristic approaches, including genetic algorithms~\cite{guidotti2024gentree} and interpolation strategies~\cite{gomez2018automatic}. Unlike existing approaches for learning \textsc{dt}s through LSO~\cite{guidotti2024gentree}, our proposal leverages optimization via surrogate models, enabling the use of gradients and making exploration more efficient by avoiding expensive black-box evaluations.

\smallskip
\textit{Tree Transformers.} Although transformers were originally designed for sequential data~\cite{vaswani2017attention}, they can be easily adapted on various domains, including graphs~\cite{dong2022pace,luo2023transformers} and trees~\cite{wang2019tree}. Trees are typically represented as linear sequences of tokens obtained through depth-first or breadth-first traversals~\cite{shiv2019novel, tang2021ast}. A key challenge lies in encoding tree structure through suitable positional representations. Early approaches introduce absolute positional embeddings based on root-to-node paths~\cite{shiv2019novel}, later extended to arbitrarily deep trees and enriched with relative attention biases~\cite{peng2022rethinking}. 
Other methods combine path-based encoding with recurrent mechanisms~\cite{peng2021integrating} or design embeddings capturing depth and sibling relations coupled with attention masking~\cite{bartkowiak2025seamlessly}.
Tree Transformers have been applied across different tasks, such as code summarization~\cite{tang2021ast}, dependency parsing~\cite{zhao2024dependency}, and molecular modeling~\cite{inukai2025frattvae}. 

Following these lines of research, we propose a \textsc{t}ree \textsc{t}ransformer \textsc{v}ariational \textsc{a}uto-\textsc{e}ncoder (\ttvae{}) where \textsc{dt}s are encoded as linear sequences of tokens in depth-first order and structural information is represented with tree absolute positional embeddings from~\cite{shiv2019novel}.

\section{Methodology}
\label{sec:method}
In this section, we present \trevis{}, a method for learning \textsc{t}ree \textsc{re}presentations from \textsc{v}ariational \textsc{i}nference in latent \textsc{s}pace. 
In the following, we first describe how \textsc{dt}s are represented as linear sequences of tokens enriched with tree structural information. 
We then detail our \textsc{t}ree \textsc{t}ransformer \textsc{v}ariational \textsc{a}uto-\textsc{e}ncoder (\ttvae{}) architecture, which is used to learn a latent space of \textsc{dt}s from such representations.
Finally, we explain how the learned latent space can be navigated by optimizing a differentiable surrogate model with gradient ascent.
The overall procedure of \trevis{} is summarized in Algorithm~\ref{alg:algorithm}, which is detailed in the subsequent sections. 

\subsection{Tree Linearization}
\label{sec:linearization}

\smallskip
\noindent \textbf{Tokenization.} 
Let $(\mathcal{X}, \mathcal{Y})=\{(x_i,y_i)\}_{i=1}^n$ be a labeled training dataset, where $x_i \in \mathcal{X}\subseteq\mathbb{R}^m$ denotes an instance and $y_i\in\mathcal{Y}=\{1,\ldots,C\}$ its class label. 
Given a maximum depth $\gamma \in \mathbb{N}$, we denote by $\mathcal{T}$ the set of all binary \textsc{dt}s of depth at most $\gamma$ that can be constructed from $(\mathcal{X}, \mathcal{Y})$. Since computing $\mathcal{T}$ is intractable, different approaches might be used to identify or enumerate a constrained subset $\mathcal{T}' \subseteq \mathcal{T}$, e.g., to approximate the Rashomon set of similarly performing \textsc{dt}s~\cite{xin2022exploring,hsu2026rashomon}. We detail in Section~\ref{sec:experiments} the specific strategy adopted in our implementation to build $\mathcal{T}'$.

For any \textsc{dt} $T \in \mathcal{T}$, each internal node applies a binary split defined by a pair $(\psi,\tau)$, where $\psi$ denotes one of the $m$ features in $\mathcal{X}$ and $\tau \in \mathbb{R}$ a threshold value. 
We define an invertible encoding function $\pi: \mathcal{T} \rightarrow \mathcal{S}$ such that, given a tree $T \in \mathcal{T}$, $\pi(T)$ is a linear sequence of tokens $S \in \mathcal{S}$ obtained through a depth-first pre-order traversal of $T$. 
Specifically, each visited node $j$ is encoded as: 
(\emph{i}) a pair of consecutive tokens ($\psi_j$, $\tau_j$), where $\psi_j$ represents the feature and $\tau_j$ its threshold, if $j$ is non-terminal; 
(\emph{ii}) a special token $\langle \text{L}\rangle$, if $j$ is terminal. 

This design choice is motivated by two reasons.
First, unlike other works~\cite{pettit2025disco}, we avoid modeling $\tau_j$ as real-valued thresholds, e.g., by using linear projections in place of embeddings, as the full domain $\mathbb{R}$ would unnecessarily enlarge the search space. 
Indeed, let $v_j^{(1)} < v_j^{(2)} < \ldots < v_j^{(n)}$ denote the sorted values of a feature $\psi_j$. 
For any consecutive pair $v_j^{(l)}, v_j^{(l+1)}$, every threshold $\tau \in [v_j^{(l)}, v_j^{(l+1)})$ yields the same partition on $\mathcal{X}$. 
Second, we do not need to represent predictions at terminal nodes, as they can be inferred with the majority class of the examples reaching that node.
As a consequence of the first point, to provide a complete set of thresholds representing all possible partitions on $\mathcal{X}$, it is sufficient to consider only one canonical value for each possible interval $[v^{(l)}, v^{(l+1)})$. 
While traditional algorithms select the midpoints between consecutive values~\cite{breiman1984classification}, in our setting this is impractical, as it would significantly increase the size of the vocabulary. Instead, we select as canonical value the left endpoint of such interval ($v^{(l)}$), so that threshold tokens in the vocabulary are restricted to values in $\mathcal{X}$. To further reduce the vocabulary size, we round such values to a pre-defined floating precision.
\begin{algorithm}[t]
\caption{\trevis{}$(\mathcal{T}, \lambda, n')$}
\label{alg:algorithm}
    \begin{algorithmic}[1]

    \Require
    $\mathcal{T}$ set training trees,  $\lambda$ sparsity regularization,
    $n'$ number of latent samples,
    $\eta$ learning rate
    
    \Ensure
    $T^\ast$ optimized decision tree
    
    \State $\mathcal{S} \gets \{\pi(T) | T \in \mathcal{T}\}$ \Comment{linearize trees}\label{line:linearize}
    
    \State $Q_\phi(z| \pi(T)), P_\theta(\pi(T) | z) \gets \text{\textit{fit\_ttvae}}(\mathcal{S})$ \Comment{fit \ttvae{}}\label{line:fit_vae}
    
    \State $\mathcal{Z} \gets \{Q_\phi(z|S) \mid S \in \mathcal{S})\}$ \Comment{get train latent}\label{line:get_z}
    
    \State $r \gets \{ J_\lambda(T) \mid T \in \mathcal{T}\}$ \Comment{evaluate fitness}\label{line:eval_fitness1}
    
    \State $g \gets \text{\textit{fit\_surrogate}}(\mathcal{Z}, r)$ \Comment{fit surrogate}\label{line:fit_sur}
    \State $\mathcal{Z}' \gets \{z_i \sim \mathcal{N}(0, I) \mid \forall i=1,\dots, n'\} $ \Comment{sample latent}\label{line:sample_z_opt}
    \For{$z \in \mathcal{Z}'$}\Comment{for each latent}\label{line:start_grad}
    \Repeat\Comment{iterate}
        \State $z \gets z + \eta \nabla_z g(z)$
        \Comment{gradient ascent}
    \Until{\textit{stop condition}}\Comment{terminate}
    \EndFor\label{line:end_grad}
    
    \State $\hat{\mathcal{T}} \gets \{\pi^{-1}(P_\theta(z))\mid z \in \mathcal{Z}'\}$ \Comment{decode trees}\label{line:decode_back}
    
    \State $T^\ast \gets \operatorname*{arg\,max}_{T \in \hat{\mathcal{T}}} J_\lambda(T) $ \Comment{select best}\label{line:start_select}
    \State \Return $T^\ast$\label{line:end_select}
    \end{algorithmic}
\end{algorithm}
\begin{figure}[t]
    \centering
    \includegraphics[width=0.5\linewidth, trim={0cm 3.5cm 0cm 1cm}, clip]{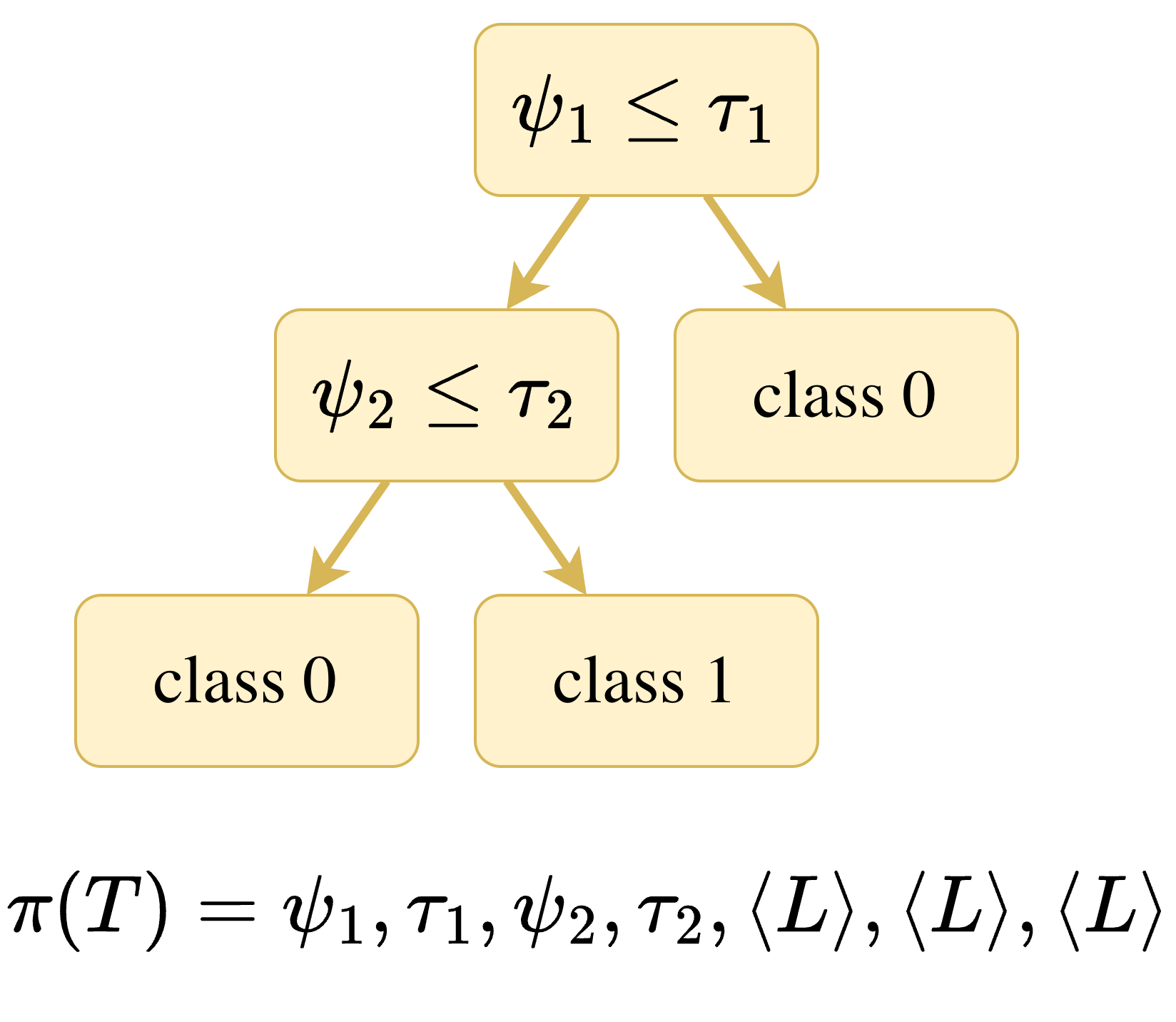}
    \caption{Example of \textsc{dt} tokenization. The tree is traversed in depth-first pre-order. Each non-terminal node is represented with two consecutive tokens, one for the selected feature ($\psi_j$) and one for its threshold ($\tau_j$). Each terminal node is represented with a special token $\langle L\rangle$.}
    \label{fig:linearize_example}
\end{figure}
Overall, our vocabulary of tokens can be restricted to: 
(\emph{i}) feature names; 
(\emph{ii}) all the distinct values in $\mathcal{X}$, except for the maximum value of each feature, approximated to a pre-defined floating precision; 
(\emph{iii}) special tokens, i.e., $\langle \text{L}\rangle$, along with those generally employed in transformer architectures: $\langle \text{CLS}\rangle$, $\langle \text{BOS}\rangle$, $\langle \text{EOS}\rangle$, $\langle \text{UNK}\rangle$~\cite{devlin2019bert, Radford2019LanguageMA}. 
Figure~\ref{fig:linearize_example} illustrates an example of tokenization of a \textsc{dt}, where $\pi$ first gathers tokens from the root ($\psi_1, \tau_1$), then from the left subtree ($\psi_2, \tau_2, \langle\text{L}\rangle, \langle\text{L}\rangle$), and finally from the right subtree ($\langle\text{L}\rangle$).

\medskip
\noindent \textbf{Positional Encoding.} 
Transformer-based architectures leverage Self-Attention (SA) to model dependencies among input tokens (Eq.~\ref{eq:sa}). 
However, SA is permutation-invariant and does not inherently encode the structure of the input. 
The most common solution to inject positional information is through absolute positional embeddings~\cite{vaswani2017attention}, added to or concatenated with token embeddings.
To properly represent the hierarchical structure of \textsc{dt}s, we add to token embeddings the absolute tree positional embeddings proposed in~\cite{shiv2019novel}, where the position of each token at node $j$ is represented by a vector of stacked one-hot chunks encoding the path from the root to $j$, with each level scaled according to a geometric series. We show the effectiveness of tree positional embeddings in Section~\ref{sec:sensitivity_analysis}.

%\smallskip
Thus, as outlined in Algorithm~\ref{alg:algorithm}, \trevis{} takes as input a set $\mathcal{T}$ of binary \textsc{dt}s with maximum depth $\gamma$ built on $(\mathcal{X}, \mathcal{Y})$ and linearize them through $\pi$ (Alg.~\ref{alg:algorithm}, line~\ref{line:linearize}). 

\subsection{Tree Transformer Variational Auto-Encoder}
\label{sec:ttvae}
Then, \trevis{} leverages our proposed \ttvae{} to learn a latent space of \textsc{dt}s in $\mathcal{T}$ (Alg.~\ref{alg:algorithm}, line~\ref{line:fit_vae}). 
VAEs generate data from a latent variable $z$ according to a conditional distribution $p(x|z)$~\cite{kingma2019introduction}. 
Since the true posterior is generally intractable, VAEs rely on variational inference and introduce an approximate posterior $q(z|x) \approx p(z|x)$. 

\begin{figure}[t]
    \centering
    \includegraphics[width=0.9\linewidth]{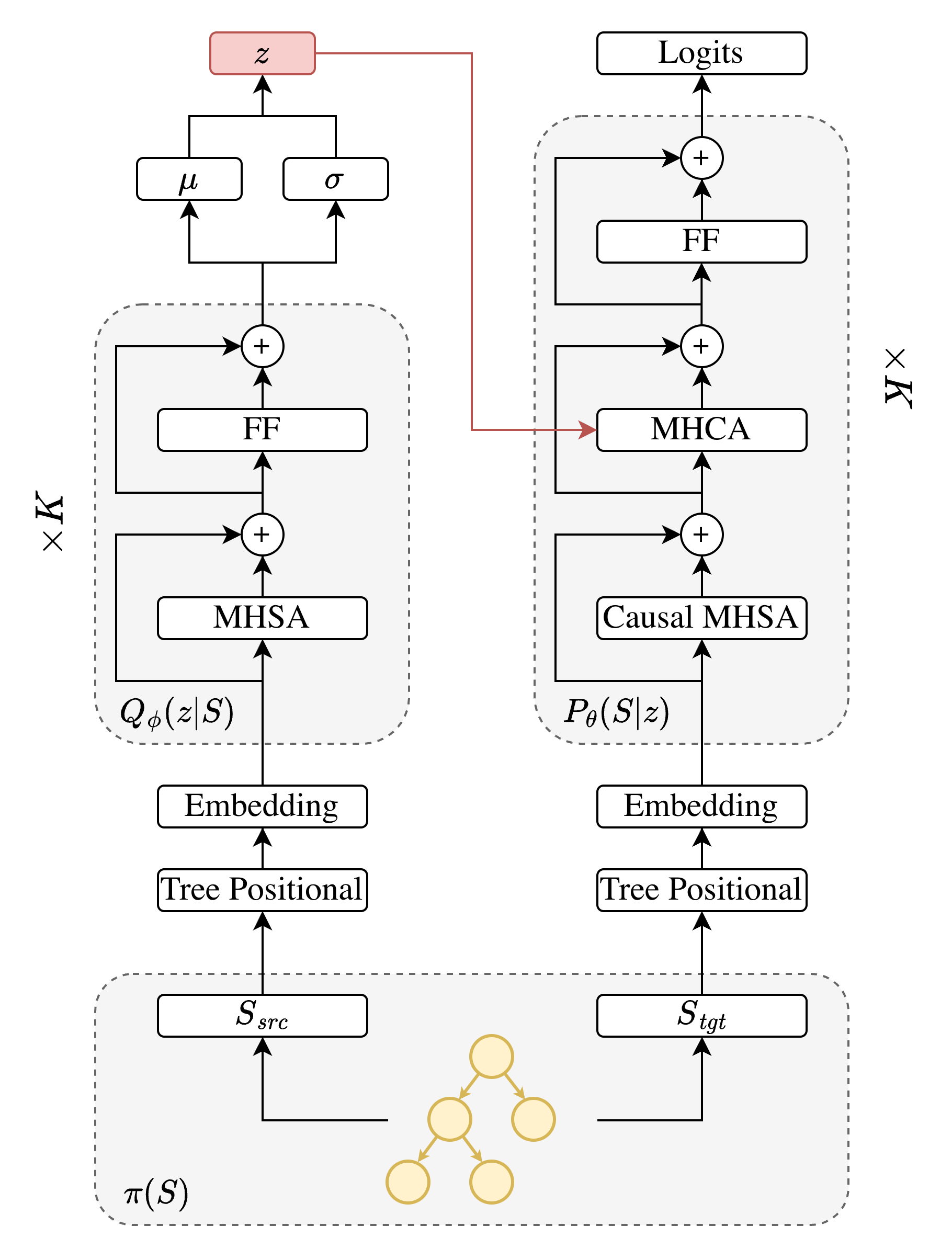}
    \caption{\ttvae{} architecture. Each training tree is linearized into a source ($S_{\textit{src}}$) and a target ($S_{\textit{tgt}}$) sequence, which are projected onto positional and token embeddings. At training time, the encoder computes $\mu$ and $\sigma$. The latent representation $z$ is sampled via the reparametrization trick and a distinct linear projection is injected into each decoder layer via MHCA. Both encoder and decoder have $K$ layers (transformer blocks).}
    \label{fig:ttvae}
\end{figure}

A full representation of the proposed \ttvae{} is provided in Figure~\ref{fig:ttvae}. 
It consists of an \textit{encoder} $Q_\phi(z|\pi(T))$ parametrized by $\phi$, modeling $p(z|\pi(T))$; and a \textit{decoder} $P_\theta(\pi(T)|z)$ parametrized by $\theta$, modeling $p(\pi(T)|z)$. 
Both components leverage the transformer architecture~\cite{vaswani2017attention}, which process the input sequence $\pi(T) = S$ through $K$ stacked \textit{transformer blocks}. Each block comprises a Multi-Head Self-Attention (MHSA) sub-layer, defined as:
\begin{align}
    &\mathit{SA}_i(S) = \mathit{Softmax}\left(\frac{(SW^i_Q) (SW^i_K)^\top}{\sqrt{d_k}}\right) SW^i_V\label{eq:sa}\\
    &\mathit{MHSA}(S) = \mathit{Concat}(\mathit{SA}_1(S),\ldots, \text{SA}_h(S))W_O,
\end{align}
where $W^i_Q \in \mathbb{R}^{k \times d_k}$, 
$W^i_K \in \mathbb{R}^{k \times d_k}$, and 
$W^i_V \in \mathbb{R}^{k \times d_v}$ are the query, key, and value 
projection matrices for the $i$-th head, respectively; 
$W_O \in \mathbb{R}^{h d_v \times k}$ is the output projection matrix; 
and $k$ is the embedding size. The MHSA sub-layer is followed by a 
Feed-Forward (FF) sub-layer, and both are equipped with residual skip connections and layer normalization.

The encoder and the decoder operate on two distinct versions of the input sequence $S$. 
For the encoder, we use $S$ prepended with the $\langle \text{CLS}\rangle$ token ($S_{\mathit{src}}$); while for the decoder, we used a shifted version of $S$ delimited by $\langle \text{BOS}\rangle$ and $\langle \text{EOS}\rangle$ tokens ($S_{\mathit{tgt}}$). 
Additionally, for the decoder the standard MHSA is replaced with \textit{Causal} MHSA, masking future positions to enforce autoregressive factorization~\cite{vaswani2017attention}. 

The encoder generates the latent mean $\mu$ and variance $\sigma$ as linear projections of the $\langle \mathrm{CLS} \rangle$ final hidden state. 
The latent representation $z$ is sampled via the reparameterization trick:
\begin{equation*}
z = \mu + \sigma \odot \epsilon,
\qquad
\epsilon \sim \mathcal{N}(0, I),
\end{equation*}
enabling the flow of gradients; then it is projected onto each decoder block and injected through a Multi-Head Cross-Attention (MHCA) sub-layer: 
\begin{align*}
    & z_l = zW_l\\
    & \text{CA}_i(H,z_l) = \text{Softmax} \left(\frac{ (HW^i_Q)(z_lW^i_K)^\top}{\sqrt{d_k}}\right) z_lW^i_V\\
    & \text{MHCA}(H,z_l) = \text{Concat} \left(\text{CA}_1(H,z_l), \ldots, \text{CA}_h(H,z_l)\right)W_O,
\end{align*}
with $H$ denoting the decoder hidden states and $W_l$ a layer-specific projection matrix.

We train the \ttvae{} on linearized \textsc{dt}s $\mathcal{S}$ (Alg.~\ref{alg:algorithm}, line~\ref{line:fit_vae}) to maximize a weighted Evidence Lower Bound ($\beta$-ELBO):
\begin{equation*}
    \mathbb{E}_{Q_\phi(z|S)} \left[ \log P_\theta(S|z) \right] - \beta \, D_{\mathrm{KL}} \left(Q_\phi(z|S) \,\|\, p(z) \right)
\end{equation*}
where $\beta \in \mathbb{R}^+$ controls the trade-off between reconstruction accuracy and latent-space regularization and can be scheduled during training to prevent the well-known problem of \textit{posterior collapse} or \textit{KL vanishing}~\cite{bowman2016generating}. We assume the prior to be $p(z) = \mathcal{N}(0, I)$, i.e., a Gaussian distribution with zero mean and identity covariance matrix.

At inference time, generation is performed autoregressively by conditioning the decoder with a latent vector $z \sim \mathcal{N}(0, I)$ and the so-far generated tokens:
%
%\begin{equation*}
%    P_\theta(S|z) = \prod_{t=1}^{l} P_\theta(S^{(t)}) \mid S^{<(t)}, z)
%\end{equation*}
%
%
\begin{equation*}
    P_\theta(S|z) = \prod_{t=1}^{l} P_\theta(S^{(t)} \mid S^{<(t)}, z)
\end{equation*}
until a termination condition is met, i.e., the generation of the $\langle \text{EOS} \rangle$ token. 
For ease of notation, we will denote autoregressive generation simply as $P_\theta(z)$. 

\begin{figure}[t]
    \centering
    \includegraphics[width=0.8\linewidth]{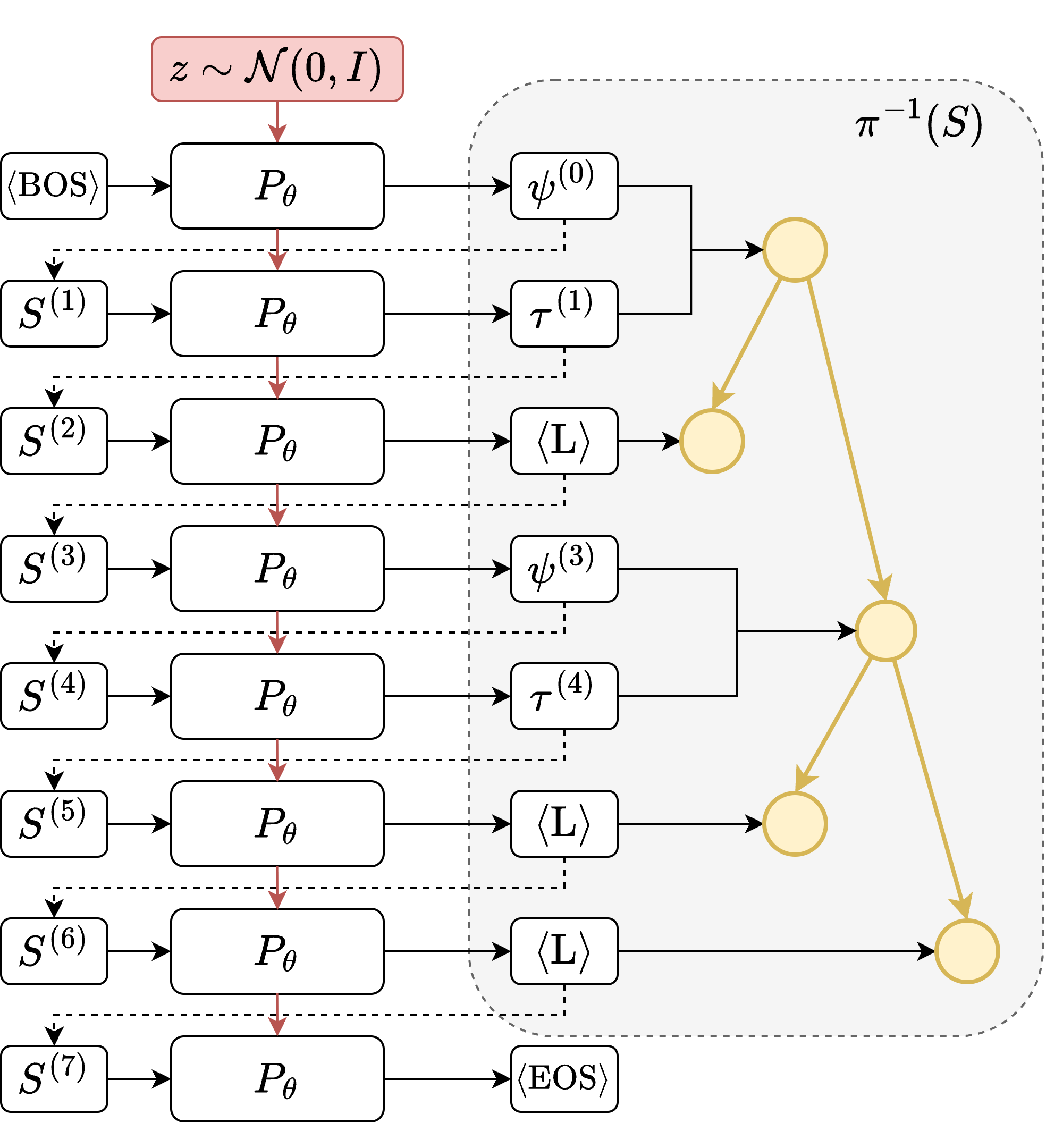}
    \caption{Autoregressive generation of a \textsc{dt}. Given a latent representation $z \sim \mathcal{N}(0, I)$ and the $\langle \text{BOS}\rangle$ token, the decoder iteratively generates tokens, appending each prediction to the input sequence until the $\langle \text{EOS}\rangle$ token is produced. 
    The generated sequence is then mapped back to a \textsc{dt} through the inverse encoding function $\pi^{-1}(S)$.}
    \label{fig:inference}
\end{figure}

In Figure~\ref{fig:inference} we show an example of autoregressive generation. 
Here, the latent vector $z$ is sampled from the prior $\mathcal{N}(0, I)$. 
Starting from the $\langle\text{BOS}\rangle$ token, at each step $i$ the decoder generates the next token, which is appended to the input $S^{(i+1)}$ for the next step, until termination, i.e., the generation of $\langle\text{EOS}\rangle$ token. 
Once generation is completed, we use the inverse encoding function $\pi^{-1}(S)$ to map the generated sequence $S$ to the correspondent \textsc{dt}.

\subsection{Latent Space Optimization via Surrogate Model}
\label{sec:optim}
After training the \ttvae{}, \trevis{} leverages its latent space $\mathcal{Z}$ to search for latent representations of \textsc{dt}s that optimize a given objective function.  
Although \trevis{} is agnostic to its optimization objective, in this work we leverage it to search for \textsc{dt}s that balance structural sparsity and predictive performance. Therefore, given a \textsc{dt} $T$ and a sparsity hyper-parameter $\lambda$, we define our optimization objective $J$ as:
\begin{equation*}
    J_\lambda(T) = \mathit{A}\!\left(T; \mathcal{X}_{\mathrm{tr}}, \mathcal{Y}_{\mathrm{tr}}\right) -\lambda V(T),
\end{equation*}
where we assume $\mathit{A}\!\left(T; \mathcal{X}_{\mathrm{tr}}, \mathcal{Y}_{\mathrm{tr}}\right)$ to be the weighted F1-score of $T$ on training data $(\mathcal{X}_{\mathrm{tr}}, \mathcal{Y}_{\mathrm{tr}})$; while $V(T)$ is the number of leaves of $T$ and $\lambda$ controls the strength of the sparsity penalty. 
While the objective is defined on trees, search is performed in the latent space. 
For a latent vector $z$, the corresponding tree is $P_{\theta}(z)$, and its objective value is $J_\lambda(P_{\theta}(z))$. We evidence that $J_\lambda(P_{\theta}(z))$ is not differentiable. A possible solution is to leverage black-box optimization algorithms~\cite{guidotti2024gentree}, which, however, are sample-inefficient. 
In contrast, we use a surrogate differentiable model $g:\mathcal{Z}\rightarrow\mathbb{R}$ to approximate $J_\lambda(P_{\theta}(z))$, e.g., a Multi-Layered Perceptron (MLP) or a linear model. 
This allows us to perform gradient-based optimization and improve the efficiency of latent-space exploration.
The surrogate model is trained on the latent representations of the training trees $\mathcal{Z}$ to predict their correspondent objective value as response $r$ (Alg.~\ref{alg:algorithm}, lines~\ref{line:get_z}-\ref{line:fit_sur}).
Given a set of $n'$ initial latent solutions $\mathcal{Z}' = \{z_i \sim \mathcal{N}(0, I) \mid \forall i=1,\dots, n'\}$ (Alg.~\ref{alg:algorithm}, line~\ref{line:sample_z_opt}), we exploit the gradient of $g$ to identify regions of the latent space that maximize the objective.
This is achieved by iteratively updating each latent representation according to 
$z \gets z + \eta \nabla_z g\!\left(z\right)$ (see Alg.~\ref{alg:algorithm}, lines~\ref{line:start_grad}--\ref{line:end_grad}), where $\eta$ is the learning rate.
The final optimized latent vectors with the
highest surrogate-predicted objective values are then decoded into \textsc{dt}
candidates and evaluated using the true objective $J_\lambda(P_\theta(z))$ (Alg.~\ref{alg:algorithm}, line~\ref{line:decode_back}).
Among the valid decoded candidates, we select the \textsc{dt} with the highest true objective value (Alg.~\ref{alg:algorithm}, line~\ref{line:start_select}).

\section{Experiments}
\label{sec:experiments}

We experiment with \trevis{} to evaluate its ability to optimize the performance-sparsity trade-off of \textsc{dt}s. In the following, we describe the experimental setting (Section~\ref{sec:experimental_setting}) and compare \trevis{} against existing \textsc{dt} learning algorithms (Section~\ref{sec:experimental_results}). We also analyze the quality of the latent space (Section~\ref{sec:latent_space_analysis}) and report a sensitivity analysis with measures for evaluating \ttvae{} generation (Section~\ref{sec:sensitivity_analysis})\footnote{The code and datasets sources are available at~\url{https://github.com/gfidone/TREVIS}. Experiments were run on a machine equipped with two AMD EPYC 9754 128-core CPUs and one NVIDIA H100 NVL GPU.}. 

\subsection{Experimental Setting}
\label{sec:experimental_setting}

\smallskip
\noindent \textbf{Datasets.} 
We evaluate \trevis{} on 18 benchmark datasets selected to cover a diverse range of sample sizes, feature types and numbers of classes. After removing instances with missing values, each dataset is split into stratified training and test sets, denoted by $(\mathcal{X}_{\mathrm{tr}},\mathcal{Y}_{\mathrm{tr}})$ and $(\mathcal{X}_{\mathrm{ts}},\mathcal{Y}_{\mathrm{ts}})$, using an 80/20\% split. We further reserve 10\% of the training data as a stratified validation set, $(\mathcal{X}_{\mathrm{vl}},\mathcal{Y}_{\mathrm{vl}})$, used for hyperparameter selection. Categorical features are one-hot encoded, and all features are min-max scaled to the range $[0,1]$.
In addition to the original continuous feature space, we evaluate \trevis{} on a discretized version obtained by using the strategy outlined in~\cite{mctavish2022fast}. This discretization trades optimality w.r.t. the original feature space for a data representation that preserves theoretical and empirical guarantees relative to a reference Gradient Boosting Decision Tree ensemble (\textsc{gbdt}). 

In this way, we reduce the potentially large search space by restricting candidate splits to those that are likely to be more informative, while also lowering the cost of \trevis{} due to a smaller vocabulary size. 
In Table~\ref{tab:datasets}, we summarize datasets information and report test performance of the \textsc{gbdt}.

\smallskip
\noindent \textbf{Tree Datasets.} 
For each training set $(\mathcal{X}_{\mathrm{tr}},\mathcal{Y}_{\mathrm{tr}})$ and for both its original and discretized version, we build four disjoint collections of \textsc{dt}s, denoted by $\mathcal{T}_{\mathrm{tr}}$, $\mathcal{T}_{\mathrm{vl}}$, $\mathcal{T}_{\mathrm{ts}}$, and $\mathcal{T}_{\mathrm{es}}$, which are used for training, validation, testing, and early stopping of \ttvae{}, respectively. 
As anticipated in Section~\ref{sec:method}, these are subsets of $\mathcal{T}$. In our implementation, each of them includes \textsc{dt}s built using randomized splits over the admissible feature-threshold pairs of $(\mathcal{X}_{\mathrm{tr}}, \mathcal{Y}_{\mathrm{tr}})$ with random depth $\gamma \in [1, 5]$ to restrict search over \textsc{dt}s whose complexity remains within an interpretable range~\cite{huysmans2011empirical}. 
Each collection contains $20{,}000$ \textsc{dt}s. As shown in Section~\ref{sec:sensitivity_analysis}, this $\mathcal{T}_{\mathrm{tr}}$ size is sufficient for \ttvae{} to achieve strong generation quality.

\begin{table}[t]
\centering
\footnotesize
\setlength{\tabcolsep}{4pt}
\caption{Dataset summary: instances ($n$), features ($m$), categorical one-hot features (${cat}^*$), discretized features (${cat}^*$), classes ($y$), class imbalance, and reference \textsc{gbdt} weighted F1-score.}
\label{tab:datasets}
\begin{tabular}{ccccccccc}
\toprule
\texttt{dataset} & $n$ & $m$ & $cat$ & ${cat}^*$ & $y$ & $\mathrm{maj}(\%)$ & $\mathrm{min}(\%)$ & \textsc{gbdt}$_{F1}$ \\
\midrule
\texttt{adult} & 12,211 & 90 & 84 & 44 & 2 & 76.07 & 23.93 & .854 \\
\texttt{bank} & 45,211 & 36 & 30 & 48 & 2 & 88.30 & 11.70 & .841 \\
\texttt{breast} & 683 & 9 & 0 & 24 & 2 & 65.01 & 34.99 & .966 \\
\texttt{compas} & 4,534 & 14 & 7 & 57 & 2 & 83.08 & 16.92 & .790 \\
\texttt{contr} & 1,473 & 9 & 3 & 23 & 2 & 57.30 & 42.70 & .731 \\
\texttt{elect} & 45,312 & 8 & 0 & 52 & 2 & 57.55 & 42.45 & .784 \\
\texttt{german} & 1,000 & 29 & 12 & 39 & 2 & 70.00 & 30.00 & .775 \\
\texttt{heart} & 261 & 15 & 9 & 21 & 2 & 62.45 & 37.55 & .856 \\
\texttt{heloc} & 10,459 & 23 & 0 & 55 & 2 & 52.19 & 47.81 & .712 \\
\texttt{iris} & 150 & 4 & 0 & 9 & 3 & 33.33 & 33.33 & .935 \\
\texttt{lrs} & 531 & 100 & 0 & 68 & 9 & 51.98 & 0.38 & .876 \\
\texttt{magic} & 19,020 & 10 & 0 & 106 & 2 & 64.84 & 35.16 & .848 \\
\texttt{pol} & 15,000 & 48 & 0 & 27 & 2 & 66.39 & 33.61 & .937 \\
\texttt{sonar} & 208 & 60 & 0 & 29 & 2 & 53.37 & 46.63 & .768 \\
\texttt{spam} & 4,601 & 57 & 0 & 62 & 2 & 60.60 & 39.40 & .922 \\
\texttt{steel} & 1,941 & 27 & 0 & 64 & 2 & 91.86 & 8.14 & .923 \\
\texttt{stud} & 1,000 & 11 & 8 & 34 & 2 & 69.60 & 30.40 & .877 \\
\texttt{wine} & 6,497 & 12 & 0 & 84 & 2 & 75.39 & 24.61 & .991 \\
\bottomrule
\end{tabular}
\end{table}

\begin{table*}[t]
\centering
\setlength{\tabcolsep}{1mm}
\caption{Weighted F1-score and number of leaves across datasets and methods. For F1-score, best values are in bold and second-best values are in italics. For leaves, lowest values are in bold and second-lowest values are in italics.}
\label{tab:f1_and_leaves_results}
\begin{tabular}{lcccccccc|cccccccc}
\toprule
 & \multicolumn{8}{c}{Weighted F1-score $\uparrow$} & \multicolumn{8}{c}{Number of Leaves $\downarrow$} \\
\cmidrule(lr){2-9} \cmidrule(lr){10-17}
\texttt{} 
& \textsc{trevis}$_{c}$ & \textsc{cart}$_{c}$ & \textsc{trevis}$_{d}$ & \textsc{cart}$_{d}$ & \textsc{dl8.5} & \textsc{dl8.5}$_{lb}$ & \textsc{gosdt}$_{lb}$ & \textsc{flow}
& \textsc{trevis}$_{c}$ & \textsc{cart}$_{c}$ & \textsc{trevis}$_{d}$ & \textsc{cart}$_{d}$ & \textsc{dl8.5} & \textsc{dl8.5}$_{lb}$ & \textsc{gosdt}$_{lb}$ & \textsc{flow} \\
\midrule
\texttt{adult} & .814 & .847 & .836 & .847 & \textbf{.859} & .851 & \textit{.855} & .746 & \textit{5} & 13 & \textbf{4} & 14 & 32 & 30 & 12 & 11 \\
\texttt{bank} & .841 & .850 & \textit{.851} & .850 & \textbf{.859} & .840 & .844 & .828 & \textit{6} & 24 & 11 & 14 & 15 & 10 & 10 & \textbf{1} \\
\texttt{breast} & .942 & .944 & .964 & .944 & \textit{.965} & .934 & \textbf{.965} & .956 & 9 & \textbf{4} & 7 & 7 & \textit{6} & 9 & \textbf{4} & 12 \\
\texttt{compas} & .770 & \textit{.799} & .795 & .789 & \textbf{.805} & .791 & .792 & .772 & 8 & \textit{5} & 13 & 8 & 16 & 23 & 20 & \textbf{2} \\
\texttt{contr} & .666 & .702 & .696 & \textit{.707} & .677 & .672 & .677 & \textbf{.709} & \textbf{10} & \textit{12} & 16 & \textit{12} & 25 & 16 & 15 & 21 \\
\texttt{elec} & .712 & .753 & .772 & .752 & \textbf{.811} & .784 & \textit{.792} & .766 & \textit{6} & \textbf{5} & 7 & 7 & 24 & 16 & 12 & \textit{6} \\
\texttt{german} & .672 & .631 & .749 & .580 & \textbf{.774} & \textbf{.774} & .755 & \textit{.760} & 10 & 14 & 8 & \textbf{6} & 8 & 8 & \textit{7} & 14 \\
\texttt{heart} & \textit{.829} & .816 & .827 & \textbf{.836} & .798 & .747 & .780 & .815 & \textbf{5} & 10 & \textit{8} & 12 & 15 & 14 & 9 & 13 \\
\texttt{heloc} & .684 & .684 & .702 & .697 & .707 & \textit{.708} & \textbf{.711} & .381 & \textbf{2} & \textbf{2} & 8 & 18 & 16 & 15 & 18 & \textit{4} \\
\texttt{iris} & \textbf{.966} & \textbf{.966} & \textbf{.966} & \textbf{.966} & \textbf{.966} & \textit{.935} & \textbf{.966} & \textbf{.966} & \textbf{4} & \textbf{4} & \textbf{4} & \textbf{4} & \textbf{4} & \textit{6} & \textbf{4} & \textbf{4} \\
\texttt{lrs} & .663 & .797 & \textit{.815} & \textbf{.825} & .776 & .753 & \textit{.815} & .804 & \textbf{7} & 14 & \textit{11} & \textit{11} & 30 & 29 & 13 & 25 \\
\texttt{magic} & .741 & .819 & .792 & .821 & \textbf{.844} & \textit{.840} & .794 & .510 & \textit{5} & 17 & 10 & 18 & 29 & 26 & \textit{5} & \textbf{1} \\
\texttt{pol} & .897 & .947 & .937 & .945 & \textbf{.960} & .934 & .950 & \textit{.958} & \textbf{4} & 14 & 10 & 11 & 21 & 25 & \textit{8} & 14 \\
\texttt{sonar} & .798 & .772 & \textbf{.810} & .764 & \textit{.802} & \textit{.802} & .779 & .685 & \textbf{5} & 14 & 9 & 14 & 15 & 15 & 14 & \textit{8} \\
\texttt{spam} & .808 & .898 & .873 & .891 & \textbf{.923} & .897 & \textit{.907} & .842 & \textbf{3} & 21 & 10 & 23 & 32 & 14 & 14 & \textit{9} \\
\texttt{steel} & .898 & .905 & .906 & .915 & \textbf{.923} & \textit{.916} & \textit{.916} & .903 & \textbf{6} & \textit{7} & 8 & 12 & 14 & 15 & 18 & 23 \\
\texttt{stud} & \textbf{.888} & .862 & .864 & \textit{.867} & \textit{.867} & .846 & .855 & .864 & \textbf{4} & 12 & 9 & 12 & 27 & 29 & 10 & \textit{8} \\
\texttt{wine} & .943 & .983 & .976 & .982 & \textbf{.993} & \textit{.984} & .983 & .979 & \textbf{4} & 17 & 8 & 11 & 20 & 14 & \textit{6} & \textit{6} \\
\midrule
\texttt{avg} & .807 & .832 & \textit{.841} & .832 & \textbf{.851} & .834 & \textit{.841} & .791 & \textbf{5.72} & 11.61 & \textit{8.94} & 11.89 & 19.39 & 17.44 & 11.06 & 10.11 \\
\texttt{std} & .102 & .100 & \textbf{.085} & {.105} & {.091} & \textit{.087} & {.091} & {.156} & \textbf{2.30} & 6.16 & \textit{2.86} & 4.69 & 8.72 & 7.63 & 4.9 & 7.28 \\
\bottomrule
\end{tabular}
\end{table*}

\smallskip
\noindent \textbf{Model Configurations.} 
\label{sec:ttvae_config}
As \ttvae{} architecture we use $K = 2$ blocks for both encoder and decoder. We set the embedding size as $k = 120$. Both MHSA and MHCA sub-layers use $H = 2$ heads. We set the FF with two layers of size $4\times k$ with ReLU activation. We set to $k/2 = 60$ the dimensionality of the latent space, i.e., $\mu, \sigma$ and $z$. This configuration is motivated by the results of our sensitivity analysis (Section~\ref{sec:sensitivity_analysis}).
All \ttvae{} instances are trained with learning rate $10^{-3}$ for at most $50$ epochs, using early stopping on the $\beta$-ELBO loss computed on $\mathcal{T}_{\mathrm{es}}$, with patience $1$ and minimum improvement $0.1$. 
To prevent KL collapse, $\beta$ is set to zero for the first $10$ epochs and then linearly increased until convergence, following prior work~\cite{LiuL19a}. We also employ free bits~\cite{kingma2016freebits} to lower-bound the KL contribution of each latent dimension. 

We implement the surrogate regressor $g$ as a MLP with a $128$-sized fully-connected hidden layer and $\tanh$ activation. 
The MLP is trained with MSE loss using learning rate $10^{-3}$, weight decay $10^{-5}$ and $100$ training epochs.
Then, we set to $n'=50{,}000$ the size of candidate latent samples $\mathcal{Z}'$ (Alg.~\ref{alg:algorithm}, line~\ref{line:sample_z_opt}).
Each candidate is then optimized for $10$ gradient ascent steps (Alg.~\ref{alg:algorithm}, lines~\ref{line:start_grad}-\ref{line:end_grad}) with learning rate $\eta=10^{-3}$. 

\smallskip
\noindent \textbf{Tree Selection.} To simplify the selection of the best \textsc{dt}, we only decode the top-$500$ latent candidates $z \in \mathcal{Z}'$ with highest surrogate scores (Alg.~\ref{alg:algorithm}, line~\ref{line:decode_back}) and evaluate their true objective $J_\lambda(P_\theta(z))$ on the training set. 
The $\lambda$ hyper-parameter can be tuned to balance its contribution in $J_\lambda$. To further control sparsity, we repeat the procedure described in Alg.~\ref{alg:algorithm}, lines~\ref{line:eval_fitness1}-\ref{line:end_select} for different values of $\lambda$. Specifically, we train a separate surrogate for each $\lambda \in \{0.0, 0.0001, 0.0005, 0.001, 0.005, 0.01\}$, resulting in six independent gradient-based optimization runs. 
Finally, we select the best \textsc{dt} across candidate values of $\lambda$ as the one maximizing the weighted F1-score on $(\mathcal{X}_{\mathrm{vl}}, \mathcal{Y}_{\mathrm{vl}})$.

\subsection{Performance-Sparsity Trade-off}
\label{sec:experimental_results}

\smallskip
\noindent \textbf{Competitors.} 
We compare the best \trevis{}-optimized \textsc{dt} against greedy and near-optimal \textsc{dt} learners. 
We denote by \trevis{}$_{{c}}$ and \trevis{}$_{{d}}$ the variants trained on the original continuous and discretized feature spaces, respectively. 
Competitors include \textsc{cart} trained on the same two data representations, denoted as \textsc{cart}$_{{c}}$ and \textsc{cart}$_{{d}}$, as well as \textsc{flow}, \textsc{dl8.5}, and \textsc{gosdt}$_{\mathrm{lb}}$, which operate on discretized features only. 
\textsc{gosdt}$_{\mathrm{lb}}$ uses the lower-bound strategy of~\cite{mctavish2022fast}; we also include \textsc{dl8.5}$_{\mathrm{lb}}$, a version of \textsc{dl8.5} using the same strategy.
All near-optimal methods are run with a one-hour time limit\footnote{We use \textsc{cart} from \url{https://scikit-learn.org/stable/}~\cite{Pedregosa2011sklearn}, \textsc{dl8.5} from \url{https://github.com/aia-uclouvain/pydl8.5}~\cite{aglin2020pydl8}, \textsc{gosdt}$_{\mathrm{lb}}$ and \textsc{dl8.5}$_{\mathrm{lb}}$ from \url{https://github.com/ubc-systopia/gosdt-guesses}~\cite{mctavish2022fast}, and the single-sink \textsc{flow} implementation from \url{https://github.com/d3m-research-group/odtlearn}~\cite{vossler2023odtlearn}.}.
As done for \trevis{}, also for \textsc{gosdt}$_{\mathrm{lb}}$ we tune  $\lambda \in \{0.0001, 0.0005, 0.001, 0.005, 0.01\}$ and set the maximum depth to $\gamma{=}5$.
We exclude $\lambda{=}0.0$ for \textsc{gosdt}$_{\mathrm{lb}}$, in line with prior work~\cite{Lin2020Generalized,mctavish2022fast}, as its bound-based pruning mechanism requires a positive regularization term to avoid prohibitively expensive search.
 For \textsc{dl8.5} and \textsc{cart} models, we tune the min. leaf support over values of the form $\lceil \lambda |\mathcal{X}_{\mathrm{tr}}| \rceil$, where $|\mathcal{X}_{\mathrm{tr}}|$ is the number of training samples and $\lambda$ ranges over the same grid adopted for \trevis{}.
For \textsc{cart} models, we also tune the cost-complexity $\alpha \in \{0,0.01,0.1\}$ and maximum depth $\gamma \in \{2,3,4,5\}$. 
For \textsc{dl8.5} and \textsc{flow}, we consider $\gamma \in \{3,4,5\}$. For \textsc{flow}, we fix $\lambda{=}0.01$ due to computational constraints. 
We use the same hyperparameter ranges for models trained on continuous and discretized features.
\begin{figure}
    \centering
    \includegraphics[width=.9\linewidth, trim={0cm 0.29cm 0cm 0.25cm}, clip]{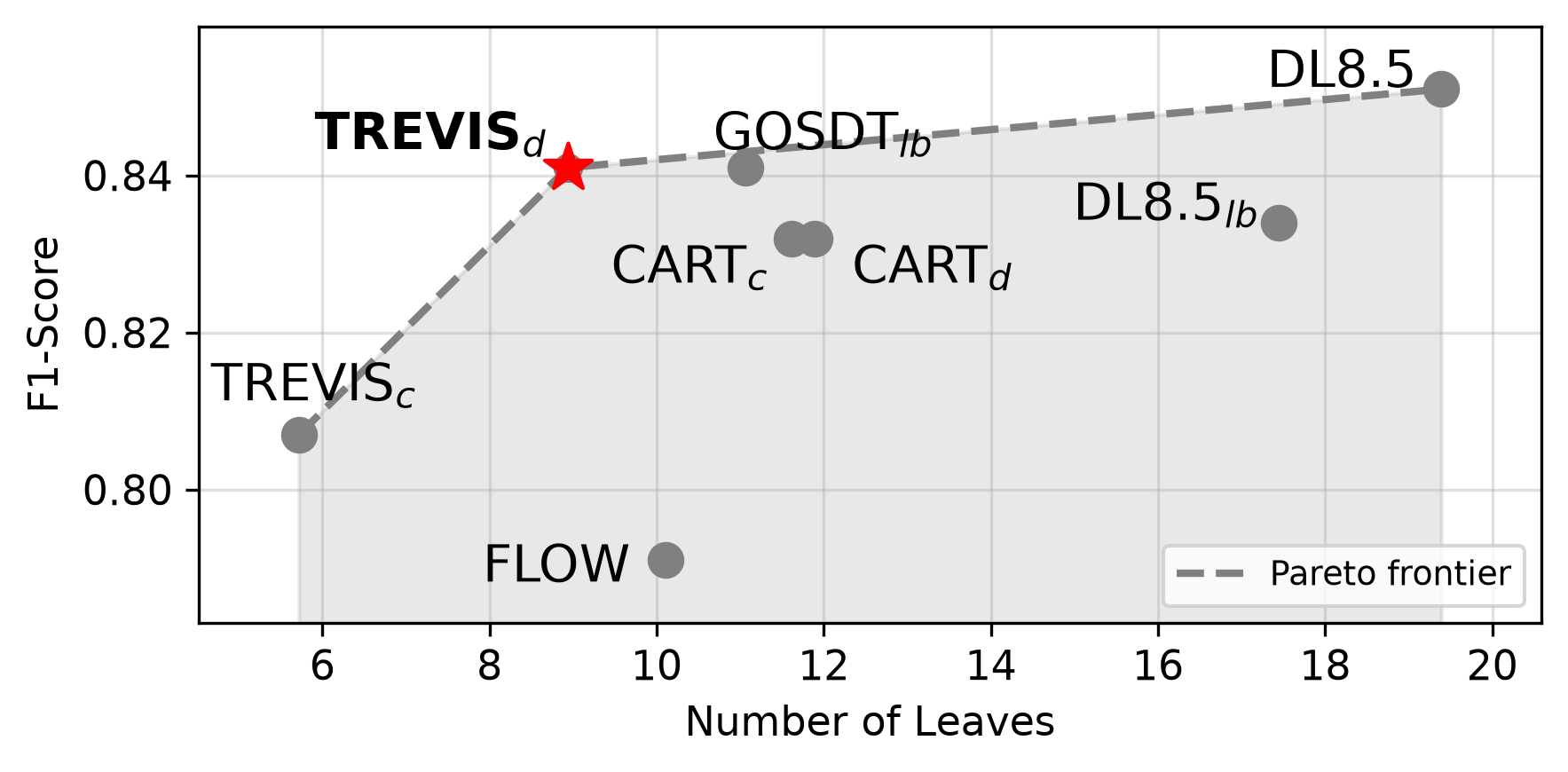}
    \caption{\textsc{dt} learning methods w.r.t. avg. weighted F1-score ($y$-axis) and number of leaves ($x$-axis). The dashed line denotes the Pareto frontier, identifying methods that jointly maximize predictive performance and minimize structural sparsity.
    Best methods are located in the upper-left corner.}
    \label{fig:f1_pareto}
\end{figure}
\begin{figure}[t]
    \centering
    \includegraphics[width=0.49\linewidth, trim={0cm 1.8cm 0cm 0cm}, clip]{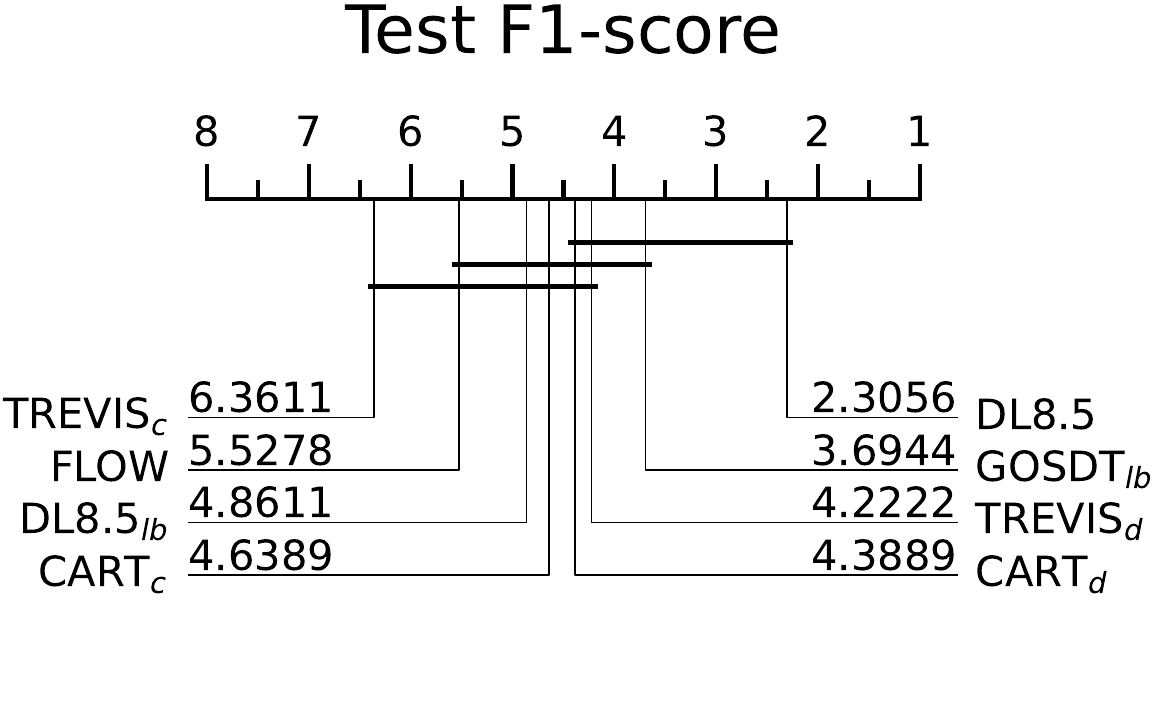}
    \includegraphics[width=0.49\linewidth, trim={0cm 1.8cm 0cm 0cm}, clip]{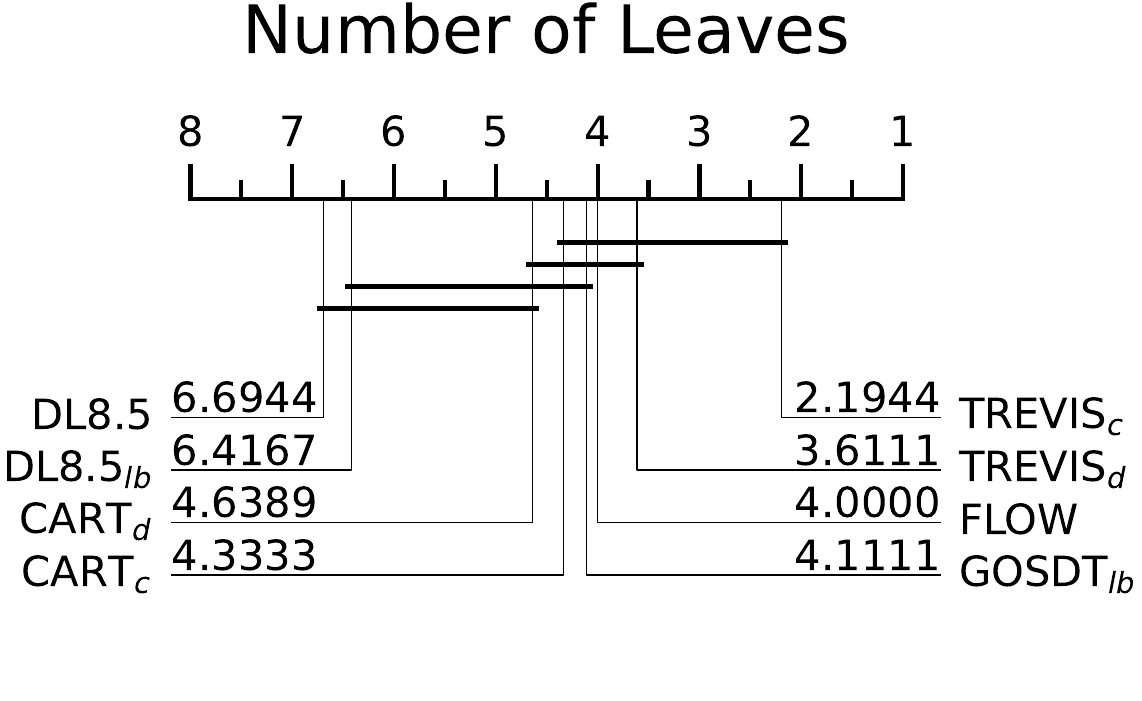}
    \caption{CD diagrams showing model rankings by test weighted F1-score (left) and number of leaves (right). Statistically indistinguishable models are connected. Best ranks on the right.}
    \label{fig:cd_test}
\end{figure}
\begin{table}[t]
    \caption{Summary of training time in seconds across datasets. For \textsc{trevis}$_{c}$/\textsc{trevis}$_{d}$, we report the avg. time of \ttvae{} training, $g$ training and gradient-based search.}
    \centering
    \scriptsize
    
    \setlength{\tabcolsep}{2mm}
    \begin{tabular}{lcccccc}
    & \multicolumn{3}{c}{\textsc{trevis}$_{c}$} 
    & \multicolumn{3}{c}{\textsc{trevis}$_{d}$} \\
    \cmidrule(lr){2-4} \cmidrule(lr){5-7}
    & \ttvae{} & $g$ training & gradient
    & \ttvae{} & $g$ training & gradient \\
    \midrule
    \texttt{avg} 
    & 360.286 & 63.138 & 199.195
    & 268.883 & 59.156 & 162.819 \\
    \texttt{std} 
    & 123.354 & 32.918 & 245.913
    & 32.495 & 39.390 & 143.942 \\
    \bottomrule
    \end{tabular}

    \begin{tabular}{ccc}
    & & \\
    \end{tabular}

    \setlength{\tabcolsep}{0.8mm}
    \begin{tabular}{lcccccccc}
    & \textsc{trevis}$_{c}$ & \textsc{cart}$_{c}$ & \textsc{trevis}$_{d}$ & \textsc{cart}$_{d}$ & \textsc{dl8.5} & \textsc{dl8.5}$_{lb}$ & \textsc{gosdt}$_{lb}$ & \textsc{flow} \\
    \midrule
    \texttt{avg} & 622.619 & 0.025 & 490.857 & 0.013 & 207.090 & 259.747 & 976.346 & 3437.446 \\
    \texttt{std} & 264.063 & 0.029 & 146.449 & 0.017 & 718.019 & 847.323 & 1234.872 & 859.180 \\
    \bottomrule
    \end{tabular}
\label{tab:training_time_summary}
\end{table}

\smallskip
\noindent \textbf{Results.}
We evaluate \trevis{} by comparing the predictive performance and structural sparsity of optimized \textsc{dt}s against competitors. 
Predictive performance is measured by the weighted F1-score computed on $(\mathcal{X}_{\text{ts}}, \mathcal{Y}_{\text{ts}})$, while sparsity is measured by the number of leaves, as in prior work~\cite{guidotti2024gentree}. 
To obtain robust estimates of the weighted F1-score, we bootstrap each test set $200$ times and report average results across resamples, following~\cite{rajkomar2018scalable}.

Table~\ref{tab:f1_and_leaves_results} reports the weighted F1-score and the number of leaves for each dataset and method, together with the corresponding average and standard deviation. 
In terms of predictive performance, \trevis{}$_{{c}}$ underperforms \textsc{cart}, while still outperforming \textsc{flow}, which reaches the time limit on most datasets\footnote{For completeness, we specify that \textsc{gosdt}$_{lb}$ exceeds the 1-hour time limit on \texttt{lrs} and \texttt{magic}, while \textsc{flow} exceeds it on all datasets except \texttt{iris}. All other competitors complete within the 1-hour limit.}. 
Conversely, \trevis{}$_{{d}}$ achieves the second-best average performance, matching \textsc{gosdt}$_{\mathrm{lb}}$ and outperforming \textsc{dl8.5}$_{{lb}}$. 
Regarding structural sparsity, both \trevis{}$_{{c}}$ and \trevis{}$_{{d}}$ consistently yield simpler \textsc{dt}s than all competing methods, yielding the lowest average number of leaves. 
This reduction is particularly evident when compared to the best-performing \textsc{dl8.5} and \textsc{dl8.5}$_{{lb}}$ approaches, which find the most complex solutions. 
Overall, \trevis{}$_{{d}}$ offers the best trade-off between predictive performance and structural sparsity\footnote{We note that surrogate models in our experiments provide reliable estimates of the objective function. 
For \textsc{trevis}$_{{d}}$, across all datasets, the MLP surrogate achieves an average test Pearson of $0.753 \pm 0.146$ and an average test RMSE of $0.027 \pm 0.018$. For \textsc{trevis}, the corresponding averages are a test Pearson correlation of $0.675 \pm 0.157$ and a test RMSE of $0.041 \pm 0.031$.}.
This is better evidenced in Figure~\ref{fig:f1_pareto}, providing a visual summary of the performance-sparsity trade-off across all methods. \trevis{}$_{{d}}$ lies in the upper-left region of the plot, indicating the most favorable balance between the two goals.

These findings are further supported by the Critical Difference (CD) plots~\cite{demvsar2006statistical} in Figure~\ref{fig:cd_test}.
Methods connected by a horizontal bar are not significantly different according to the Nemenyi test at $\alpha = 0.1$, i.e., the null hypothesis of equal average ranks cannot be rejected. 
Overall, \trevis{}$_{{d}}$ achieves a better predictive performance ranking than \textsc{cart}$_{{d}}$ and remains statistically indistinguishable from the best near-optimal learning methods, while attaining the best rank in terms of structural sparsity.

Table~\ref{tab:training_time_summary} reports the average training time of each
method, together with the individual runtimes of the main \trevis{} components.
Overall, the total runtime of \trevis{}$_{{c}}$ and \trevis{}$_{{d}}$ is
competitive with near-optimal tree-learning approaches, especially when compared
to \textsc{gosdt}$_{{lb}}$, whose runtime increases on
larger datasets. For more details about the quality of generated \textsc{dt}s under the selected \textsc{ttvae} configuration ($k=120$, $K=2$, $H=4$), see Section~\ref{sec:sensitivity_analysis}.

\begin{figure*}[t]
    \centering
    \includegraphics[width=0.9\linewidth,  trim={0cm 0.5cm 0cm 0cm}, clip]{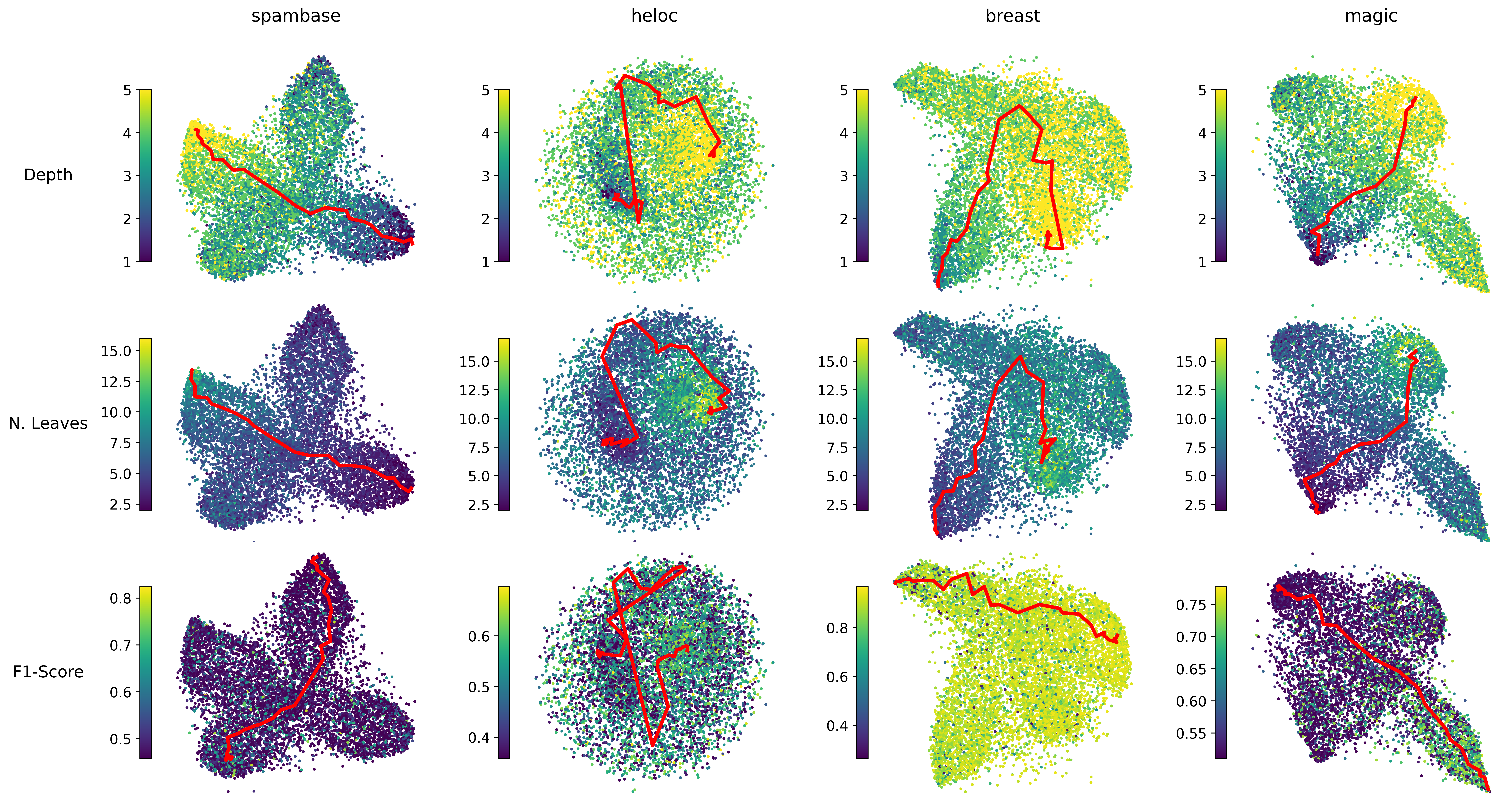}
    \caption{UMAP projections of the latent space for four datasets (columns) colored according to \textsc{dt} properties (rows). Red lines denotes trajectories obtained by moving in the direction of the gradient of the correspondent surrogate.
    }
    \label{fig:latent_spaces}
\end{figure*}

\subsection{Latent Space Analysis }
\label{sec:latent_space_analysis}
The effectiveness of \textsc{trevis} in discovering accurate \textsc{dt} representations relies on the capability of \ttvae{} to provide a structured latent space $\mathcal{Z}$ characterized by \textit{local smoothness} and \textit{global coherence}. Local smoothness ensures that the distance between two latent points $z, z' \in \mathcal{Z}$ is proportional to their distance in the output space, i.e., similar latent points correspond to similar \textsc{dt}s. Global coherence implies the existence of directions in $\mathcal{Z}$ capturing relevant semantic properties of \textsc{dt}s. Moving along these directions should yield \textsc{dt}s that change gradually in the output space w.r.t. a given property. We present here a study on local smoothness and global coherence on six representative datasets, namely \texttt{breast}, \texttt{german}, \texttt{compas}, \texttt{heloc}, \texttt{spam}, and \texttt{magic}, which vary in terms of size, class imbalance and type of features.

\smallskip
\noindent \textbf{Local smoothness}. To measure local smoothness, we sample latent points $z \sim \mathcal{N}(0, I)$ and compute perturbed counterparts as $z' = z + \epsilon \sigma$, where $\sigma \sim \mathcal{N}(0, I)$ denotes Gaussian noise and $\epsilon \sim \mathcal{U}(0, 1)$ controls the magnitude of the perturbation. Larger values of $\epsilon$ increase the distance between $z$ and $z'$. To measure the distance between the correspondent sequences $s = P_\psi(z)$ and $s' = P_\psi(z')$ in the output space, we use the normalized edit distance. Local smoothness can be estimated as the Spearman correlation $\rho$ between $\epsilon$ and edit distances.

To obtain a relative measure of local smoothness, we also employ Triplet Accuracy (TA)~\cite{WangHRS21}. Given an anchor $z_a \sim \mathcal{N}(0, I)$, we define positive and negative perturbations as $\sigma_p \sim \mathcal{N}(0, I)$ and $\sigma_n \sim \mathcal{N}(0, I)$. We define the positive example as $z_p = z_a + \epsilon \sigma_p$. To enforce $\mathit{dist}(z_p, z_a) < \mathit{dist}(z_n, z_a)$, where $z_n$ is the negative example and $\mathit{dist}$ any distance measure, we introduce a weight $\alpha > 1$ such that $z_n = z_a + \alpha\epsilon \sigma_n$. Given triplets $\mathcal{S}_{\mathit{tr}} = \{\langle s_a = P_\psi(z^i_a), s_p = P_\psi(z^i_p), s_n = P_\psi(z^i_n)\rangle\}^N_{i=1}$, we define TA as:

\begin{equation*}
    \text{TA}(\mathcal{S}_{\mathit{tr}}) = \frac{1}{|\mathcal{S}_{\mathit{tr}}|} \sum^{|\mathcal{S}_{\mathit{tr}}|}_{i=1} \mathbf{1} \left[\mathit{e}(s^i_a, s^i_p) < \mathit{e}(s^i_a, s^i_n)\right],
\end{equation*}

\noindent where $e$ is the normalized edit distance. 
Intuitively, TA measures whether local neighborhoods in the latent space are preserved after decoding: higher values indicate that nearby latent points generate more similar outputs than farther ones.
\begin{table}[t]
\centering
\setlength{\tabcolsep}{2.5pt}
\caption{Local smoothness correlation ($\rho$) and Triplet Accuracy (TA).}
\label{tab:smooth}
\begin{tabular}{ccccccccc}
%\toprule
 &
\texttt{spam} &
\texttt{german} &
\texttt{heloc} &
\texttt{breast} &
\texttt{magic} &
\texttt{compas} &
\texttt{avg} $\pm$ \texttt{std}\\
\midrule
$\rho$ & .609 & .629 & .634 & .685 & .632 & .652 & $.640\pm.240$\\
TA    & .686 & .629 & .750 & .833 & .512 & .777 & $.698\pm.105$\\
\bottomrule
\end{tabular}
\end{table}

In Table~\ref{tab:smooth} we report $\rho$ and TA ($\alpha=2.0$) computed on $1000$ latent points (triplets) across datasets. 
We observe positive correlations (all $p$-values are below $0.05$), proving that distances in the output space tend to increase consistently with distances between the corresponding latent points. This finding is further supported by positive TA scores, emphasizing how relative distances among neighboring latent points are largely preserved in the output space. 

\smallskip
\noindent \textbf{Global coherence}. We train three instances of a surrogate Lasso $g_{\mathit{lasso}}: \mathcal{Z} \rightarrow \mathbb{R}$ to separately predict the weighted F1-score, the number of leaves and the depth of a collection of \textsc{dt} latent representations $z \sim \mathcal{N}(0, I)$. For each surrogate, we generate a trajectory in the latent space by perturbing the origin $z_0 = \mathbf{0} \in \mathbb{R}^{60}$ along the direction of the normalized gradient $\nabla g_{\mathit{lasso}}(z_0)$, i.e., the weights of the linear model, by a factor $t \sim \mathcal{U}(-3, 3)$.

In Figure~\ref{fig:latent_spaces} we display UMAP projections of the sample of latent points for four datasets. 
Each column corresponds to one dataset and visualizes the same latent space colored according to the different properties of correspondent decoded \textsc{dt}s (rows). 
The red line denotes the trajectory obtained by moving from $z_0$ along the gradient direction of the corresponding surrogate model. 
The latent spaces appear smoother and more structured w.r.t. \textsc{dt} structural properties than predictive performance. 
For all properties, the trajectories identify smooth directions of variation connecting regions corresponding to the lowest and highest values. 

\subsection{Sensitivity Analysis }
\label{sec:sensitivity_analysis}
In this section we report a sensitivity analysis of different \trevis{} configurations.
In particular, we examine the impact of the shape of \ttvae{} architecture, the choice between tree and sinusoidal positional encoding, and the number $|\mathcal{T}_{\mathrm{tr}}|$ of training trees.
All analyses are conducted on the six datasets used in Section~\ref{sec:latent_space_analysis}.
As evaluation measures we consider the $\beta$-ELBO loss, including the reconstruction term and KL divergence separately. To evaluate the quality of generation, we define \textit{validity}, \textit{novelty}, and \textit{diversity}. 
Given $1000$ generated sequences, \textit{validity} is the fraction of those correctly decoded into binary \textsc{dt}s, \textit{novelty} is the fraction of valid \textsc{dt}s absent from $\mathcal{T}_{\mathrm{tr}}$ and \textit{diversity} is the average pairwise edit distance among valid sequences.

\begin{figure}[t]
    \centering
    \includegraphics[width=0.49\linewidth]{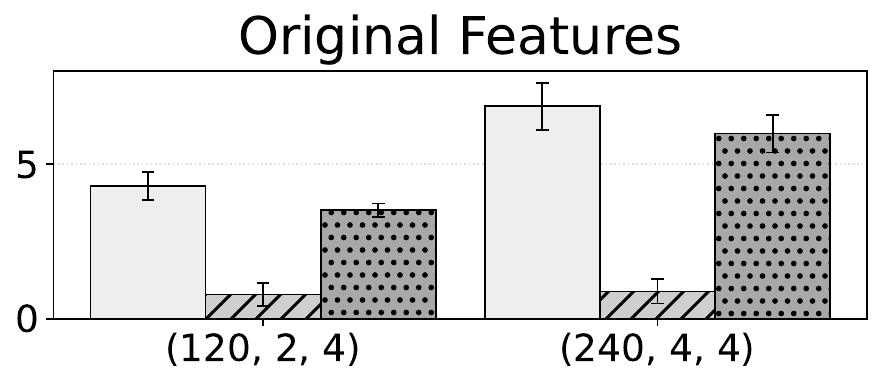}
    \includegraphics[width=0.49\linewidth]{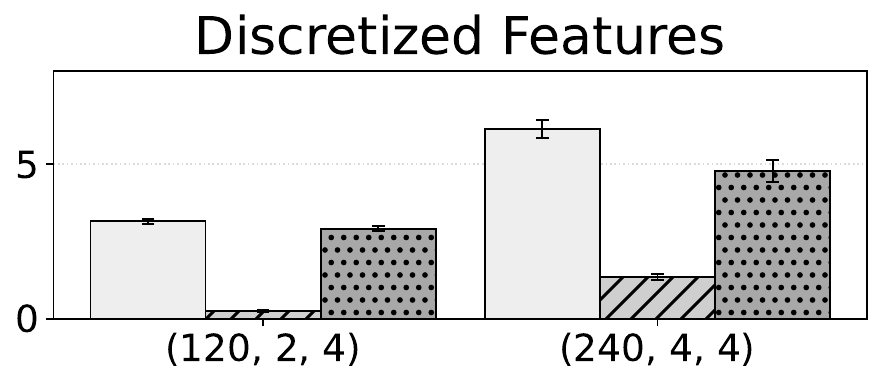} 
    \includegraphics[width=0.49\linewidth]{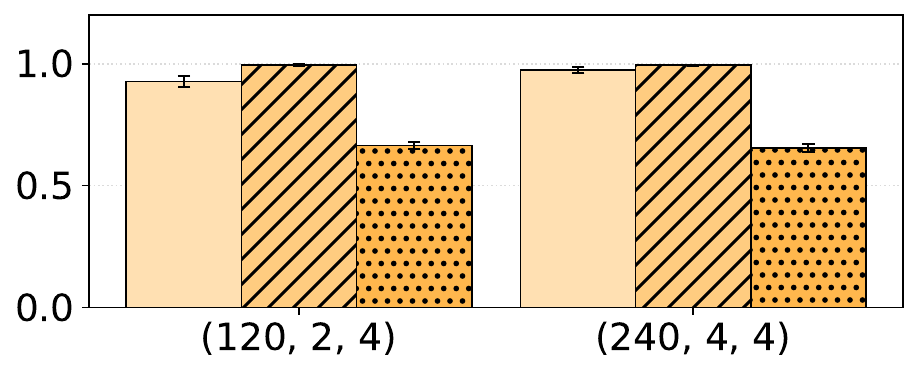} 
    \includegraphics[width=0.49\linewidth]{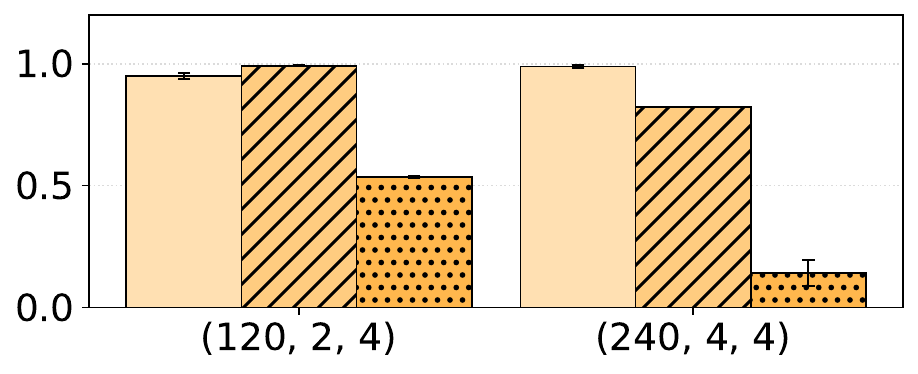} 
    \includegraphics[width=0.49\linewidth]{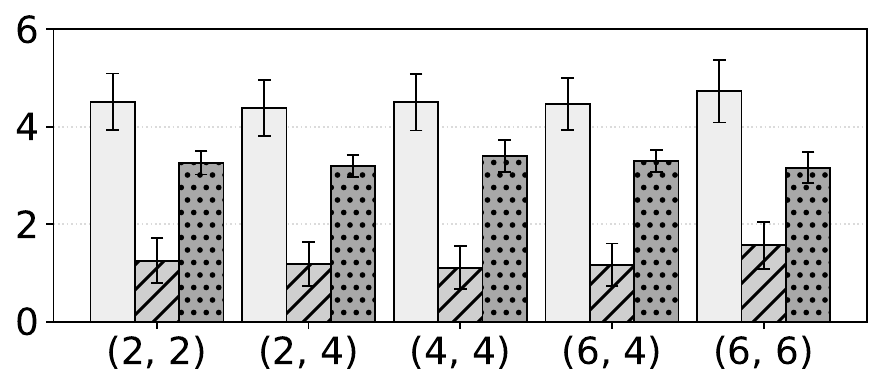}
    \includegraphics[width=0.49\linewidth]{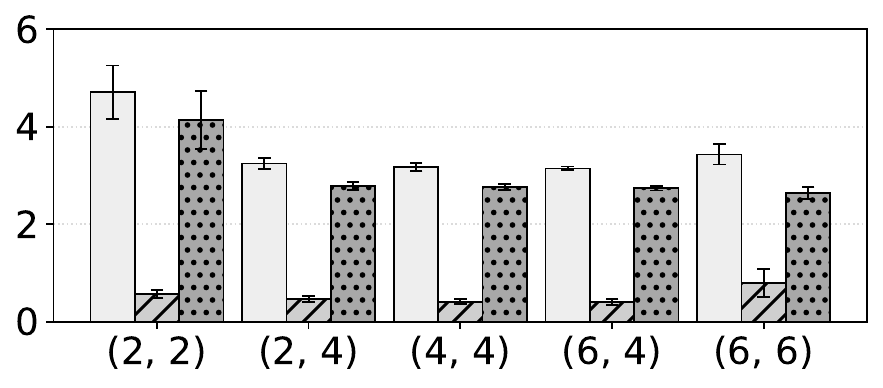}
    \includegraphics[width=0.49\linewidth]{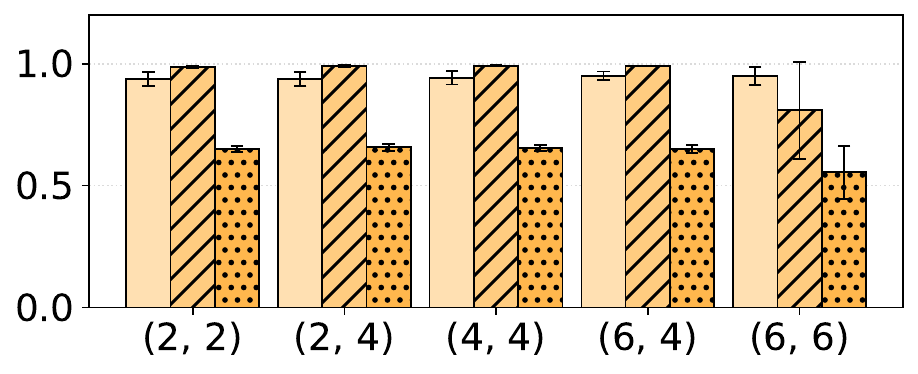} 
    \includegraphics[width=0.49\linewidth]{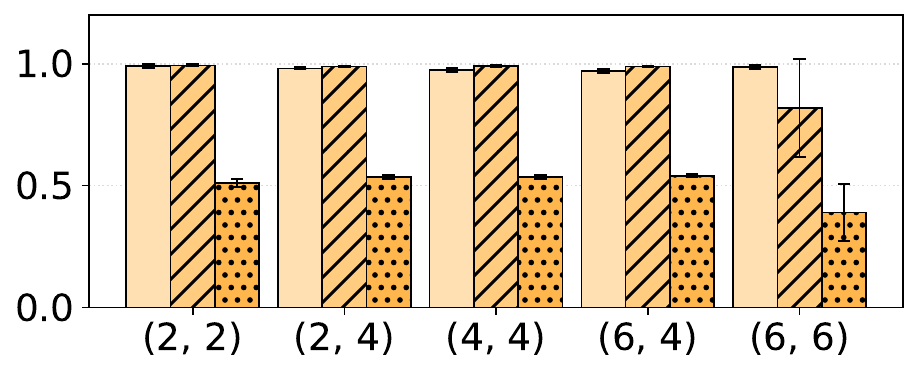} 
    \includegraphics[width=0.49\linewidth]{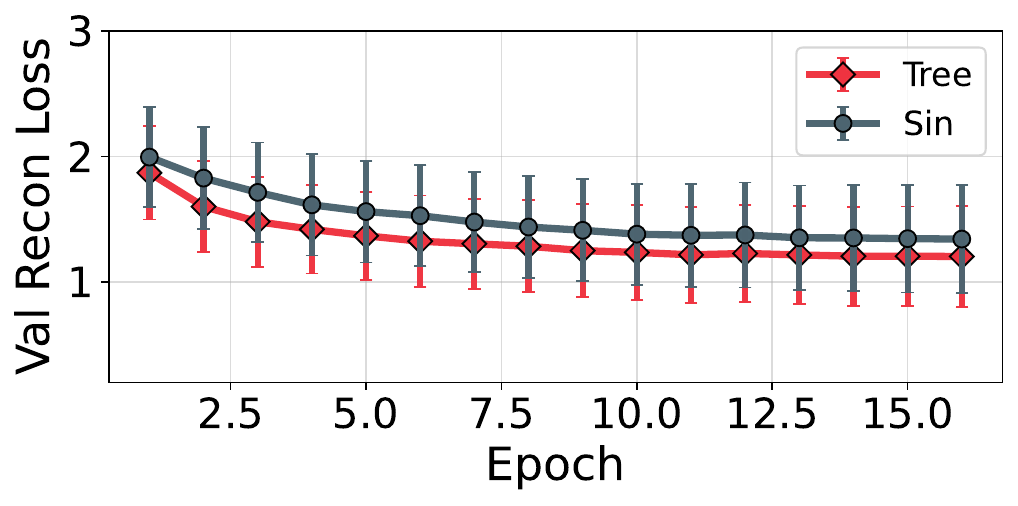}
    \includegraphics[width=0.49\linewidth]{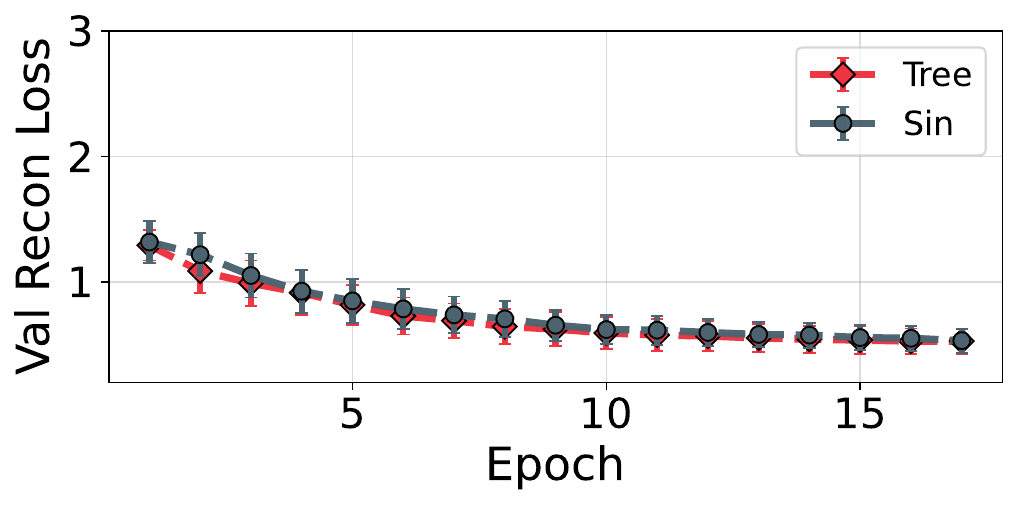}
    \includegraphics[width=1\linewidth, trim={0cm 0.9cm 0cm 0.2cm}, clip]{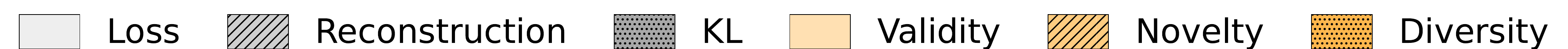}

    \caption{Effect of embedding size $k$, tree training set size $|\mathcal{T}_{\mathrm{tr}}|$, model depth $K$, attention heads $H$, and positional encoding on generative performance. Configurations are reported as $(k,K,H)$ when needed; results are averaged across datasets and shown as mean $\pm$ std.
   }
    \label{fig:sensitivity_vae}
\end{figure}

\smallskip
\noindent \textbf{Architecture and Positional encoding.} We study the impact of the \ttvae{} architecture, varying the embedding size $k$, number of layers (blocks) $K$, and number of attention heads $H$. The training hyper-parameters are kept fixed across tested configurations and match the ones described in Section~\ref{sec:ttvae_config}.

Figure~\ref{fig:sensitivity_vae} summarizes our analysis for both original and discretized feature spaces.
Regarding the embedding size, we evaluate $k \in \{120,240\}$, $K \in \{2,4\}$, and $H \in \{2,4\}$, using $|\mathcal{T}_{\mathrm{tr}}|=60{,}000$. Overall, smaller models provide better loss performance while also generating more novel and diverse pools of trees, especially in the discretized setting.
We further analyze the impact of the number of layers and attention heads by varying $K$ and $H$, while fixing $|\mathcal{T}_{\mathrm{tr}}|=20{,}000$ and $k=120$. 
The simplest architecture achieves performance comparable to more complex configurations, while substantially reducing training time. 
Across both settings, the $(120,2,2)$ configuration is faster than the larger $(120,6,4)$, reducing training time from $16.20 \pm 2.62$ to $5.89 \pm 2.77$ minutes in the discretized setting and from $9.12 \pm 1.08$ to $5.35 \pm 0.51$ minutes in the original feature space. 
Therefore, we adopted the simplest architecture for the main experiments.
In Figure~\ref{fig:sensitivity_vae} we also analyze the impact of tree and sinusoidal positional encoding on the validation reconstruction loss. On average, the tree positional encoding provides an advantage in the original feature space, achieving lower validation reconstruction loss across training epochs. These results further validate the effectiveness of the tree positional embeddings proposed by~\cite{shiv2019novel}, which we use throughout all our experiments.

\begin{figure}[t]
    \centering
    \includegraphics[scale=0.26]{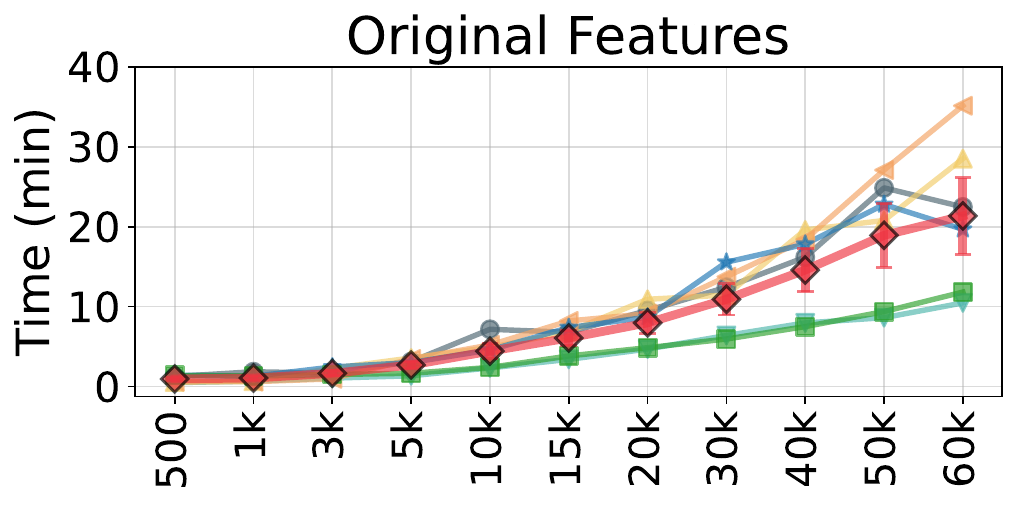}
    \includegraphics[scale=0.26, trim={1cm 0cm 0cm 0cm}, clip]{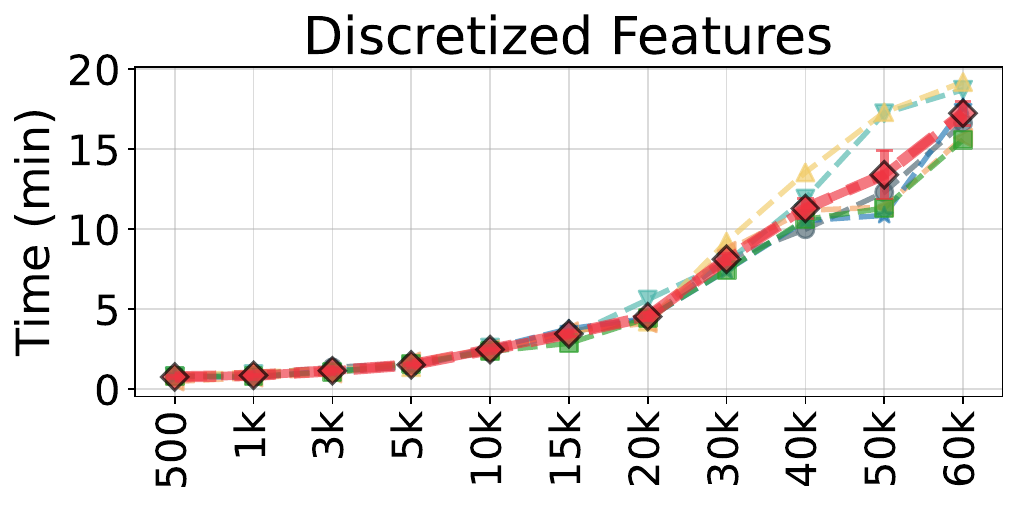}
    \includegraphics[scale=0.26]{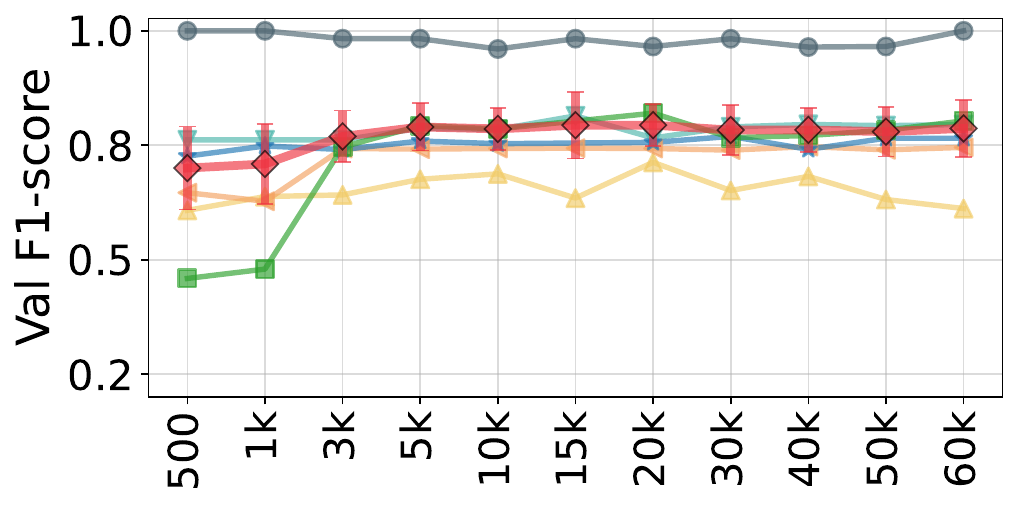}
    \includegraphics[scale=0.26, trim={1cm 0cm 0cm 0cm}, clip]{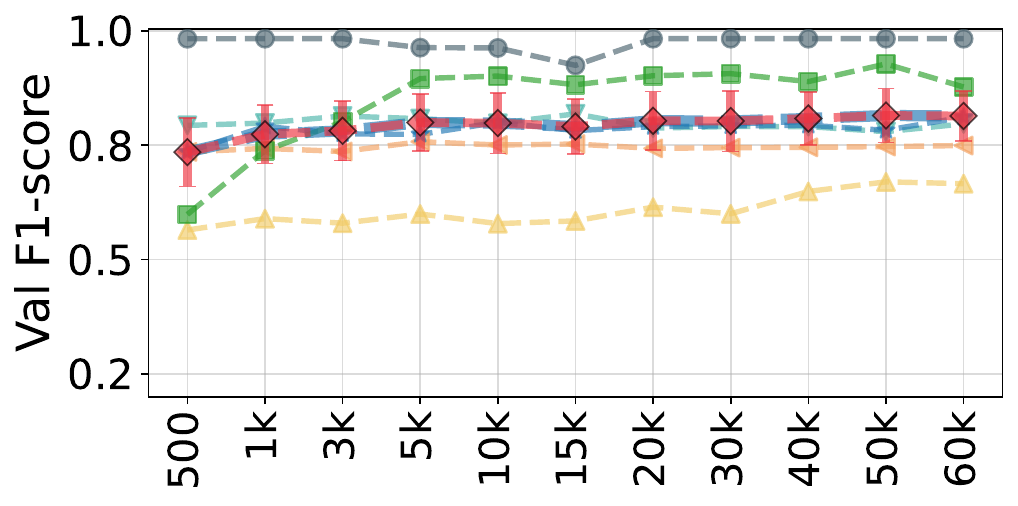}
    \includegraphics[scale=0.26]{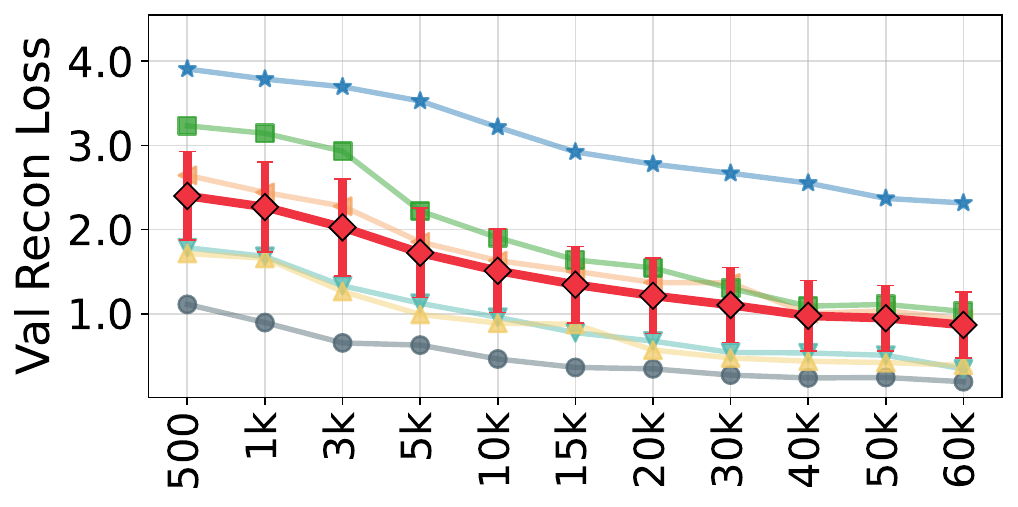}
    \includegraphics[scale=0.26, trim={1cm 0cm 0cm 0cm}, clip]{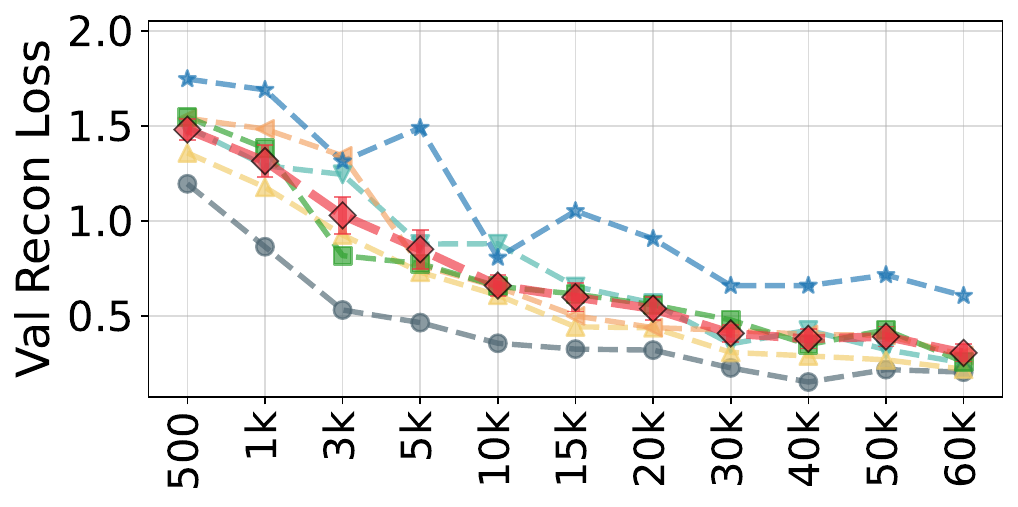}
    \includegraphics[scale=0.26]{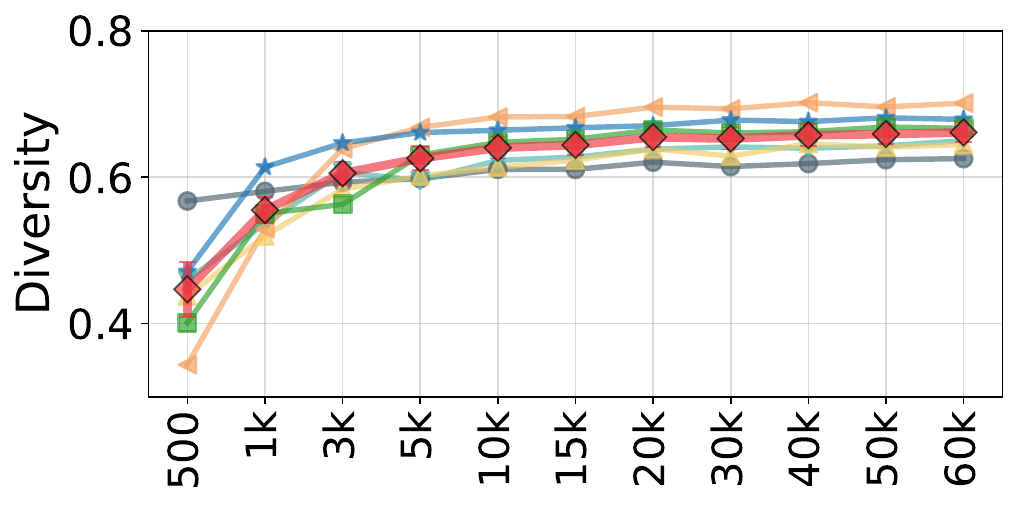}
    \includegraphics[scale=0.26, trim={1cm 0cm 0cm 0cm}, clip]{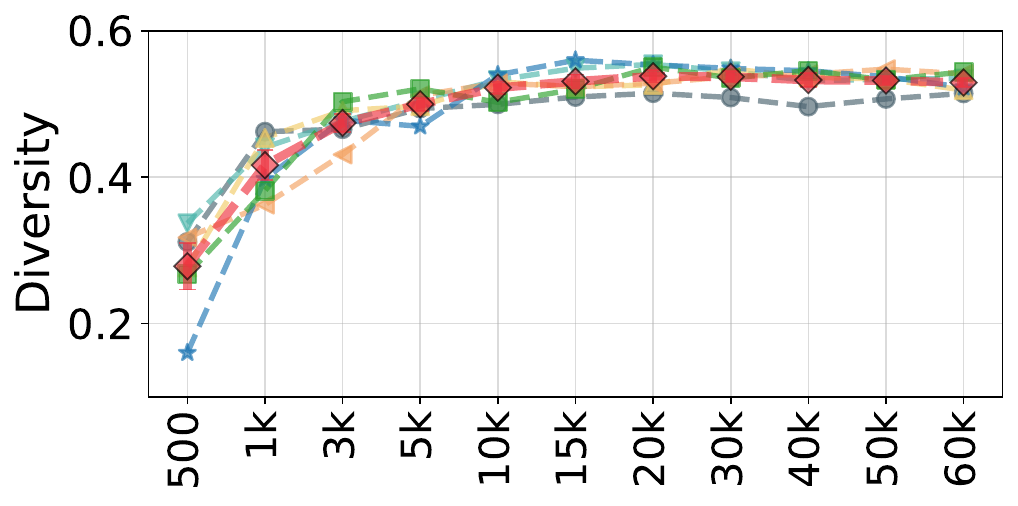}
    \includegraphics[scale=0.26, trim={0cm 1.5cm 0cm 0.5cm}, clip]{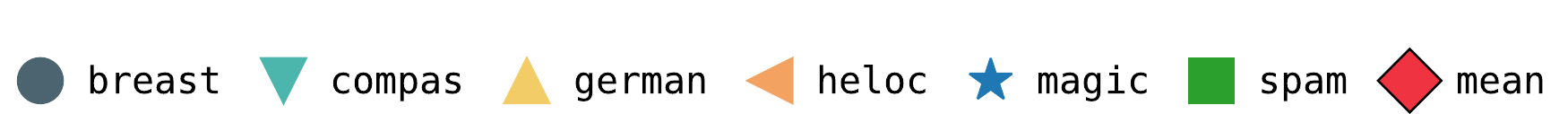}

    \caption{
    Effect of $|\mathcal{T}_\mathrm{tr}|$ on training time, weighted F1-score on $\mathcal{X}_\mathrm{vl}$, reconstruction loss on $\mathcal{T}_\mathrm{vl}$, and tree diversity.
    Solid/dashed lines denote evaluation w.r.t original/discretized features~\cite{mctavish2022fast}; error ranges show $0.5\times$ std. for visualization.
    }
    \label{fig:sensitivity_train_size}
\end{figure}

\smallskip
\noindent \textbf{Training Trees Size.}
We assess the impact of $|\mathcal{T}_{\mathrm{tr}}|$ on \ttvae{} performance and decoded-tree quality. 
Figure~\ref{fig:sensitivity_train_size} reports results for both original and discretized features. 
Training time increases almost linearly with $|\mathcal{T}_{\mathrm{tr}}|$, while the validation F1-score remains largely stable after the first few thousand trees. 
The reconstruction loss decreases as more trees are used for training, but the improvement becomes marginal beyond $20{,}000$ trees. 
Similarly, diversity increases rapidly for small training sets and then saturates. 
We therefore set $|\mathcal{T}_{\mathrm{tr}}|=20{,}000$ in the main experiments, as it offers the best trade-off between training cost, reconstruction quality, predictive performance, and diversity.

\section{Conclusion}
\label{sec:conclusion}

We have introduced \trevis{}, a framework for learning \textsc{dt}s by navigating the latent space of a \ttvae{}. By leveraging a differentiable surrogate model, \trevis{} allows to use gradient-ascent for jointly optimizing multiple \textsc{dt} properties. Our experiments show that \trevis{} discovers \textsc{dt}s with predictive performance comparable to near-optimal learning algorithms, but with improved structural sparsity, resulting in the most favorable performance-sparsity trade-off. This effectiveness can be attributed to a locally smooth latent space with directions capturing relevant \textsc{dt} properties.

Future work will extend \trevis{} beyond the performance-sparsity trade-off by considering more complex objectives including fairness, privacy, and robustness~\cite{chander2025toward}. We also plan to use \trevis{} for counterfactual exploration of \textsc{dt}s, identifying neighboring latent directions that preserve some properties while improving others~\cite{sobieski2024global}.
Finally, we aim to investigate the impact of almost-equally-optimal \textsc{dt}s~\cite{xin2022exploring} on \trevis{}, possibly uncovering alternatives trade-offs between predictive performance and other \textsc{dt} desiderata.

\trevis{} does have some limitations. Our experiments show that some \textsc{dt} properties, e.g., predictive performance, might be less organized in the latent space, making more challenging to find useful directions for optimization. 
Moreover, \trevis{} involves several training and optimization steps, including the training of the \ttvae{}, the training of the surrogate and the gradient-based optimization, making it less efficient than greedy methods. %such as \textsc{cart}.
Finally, results on the original feature space suggest that \trevis{} is less effective when considering the full set of possible splits, while substantially improving from a prior feature-selection and discretization of the search space.

%\section{Acknowledgments}
%\label{sec:ack}

\bibliographystyle{IEEEtran}
\bibliography{IEEEabrv,biblio_short_pdt}
\end{document}